\documentclass[10pt,twocolumn,letterpaper]{article}

\usepackage[pagenumbers]{cvpr} 

\usepackage{pifont}

\newcommand{\cmark}{\ding{51}}
\newcommand{\xmark}{\ding{55}}

\usepackage{colortbl}
\definecolor{oursrow}{gray}{0.92}

\usepackage[T1]{fontenc}
\usepackage{array}
\usepackage{enumitem}

\definecolor{cvprblue}{rgb}{0.21,0.49,0.74}

\usepackage[
    breaklinks,
    colorlinks=true,
    allcolors=cvprblue
]{hyperref}

\hypersetup{
    pdftitle={LeanWAM: Learning Latent World Action Models from Frozen Video Models},
    pdfauthor={Shaolong Li, Hongze Fu, Joyce Chai, Yinpei Dai},
    pdfsubject={},
    pdfkeywords={}
}

\title{LeanWAM: Learning Latent World Action Models from Frozen Video Models}

\author{
Shaolong Li$^{*}$ \qquad
Hongze Fu$^{*}$ \qquad
Joyce Chai$^{\dagger}$ \qquad
Yinpei Dai$^{\dagger}$ \\[10pt]
{\large University of Michigan}
}

\begin{document}
\maketitle

\makeatletter
\begingroup
\renewcommand{\thefootnote}{}
\let\@makefntext\@firstofone
\footnotetext[0]{%
    \noindent
    $^{*}$Equal contribution.
    \quad $^{\dagger}$Equal advising.\\
    Project website: \url{https://shaolonl-16.github.io/leanwam/}
}
\endgroup
\makeatother

\begin{abstract}
World action models (WAMs) typically adapt large pretrained video generators to jointly predict future observations and robot actions, incurring substantial training and deployment costs. However, it remains unclear which future representations a policy actually needs and how compactly they can be represented. Our probing study shows that complementary future representations can be compressed into a few tokens while retaining most of their action-decoding performance. Motivated by this finding, we introduce LeanWAM, a latent WAM that transfers video priors from a frozen video generator to a lightweight policy via two-stage training. In the first stage, a Future Extractor compresses observed transitions into 32 predictive tokens. These tokens condition the frozen generator and are jointly shaped by action, RGB, semantic, and geometric supervision, while a compact readout of the generator's hidden states supplies complementary world tokens. The second stage distills both representations into the policy, which learns to predict them from current observations while generating actions. With 1.4B parameters and no large-scale embodied pretraining, LeanWAM outperforms strong vision-language-action (VLA) and WAM baselines on LIBERO, LIBERO-Plus, and RoboTwin~2.0. It predicts an action chunk in 25.1\,ms on an A100 GPU, 12.3--41.5$\times$ faster than representative WAMs. These results demonstrate that compact predictive supervision can transfer video priors to robust and efficient control.
\end{abstract}
\section{Introduction}
\label{sec:intro}

World action models (WAMs) couple robot action learning with future prediction~\cite{dreamzero,motus,lingbotva,fastwam}. Built on pretrained video generators, they bring two benefits to control: joint video-action learning provides dense supervision for anticipating scene dynamics~\cite{uwm,dreamzero}, and video pretraining supplies spatiotemporal priors that robot demonstrations alone may not capture~\cite{cosmospolicy}. This framework also supports learning from videos without action labels~\cite{du2023learning,wu2024unleashing}.
However, these benefits come at a substantial cost. Many WAMs rely on billion-parameter video backbones over dense visual latents~\cite{motus,lingbotva,fastwam}. Their representations preserve background, texture, and appearance alongside the robot and object motion needed for control. Modeling all these details increases computation without necessarily improving actions. Compact latent actions~\cite{lapa,univla} and compressed perceptual features~\cite{rewam} offer alternatives, but do not use a pretrained video generator to shape their representations. They therefore do not directly benefit from the motion and interaction priors encoded in these generators, which could complement the supervision available in robot demonstrations.

\begin{figure}[t]
\centering
\includegraphics[width=\columnwidth]{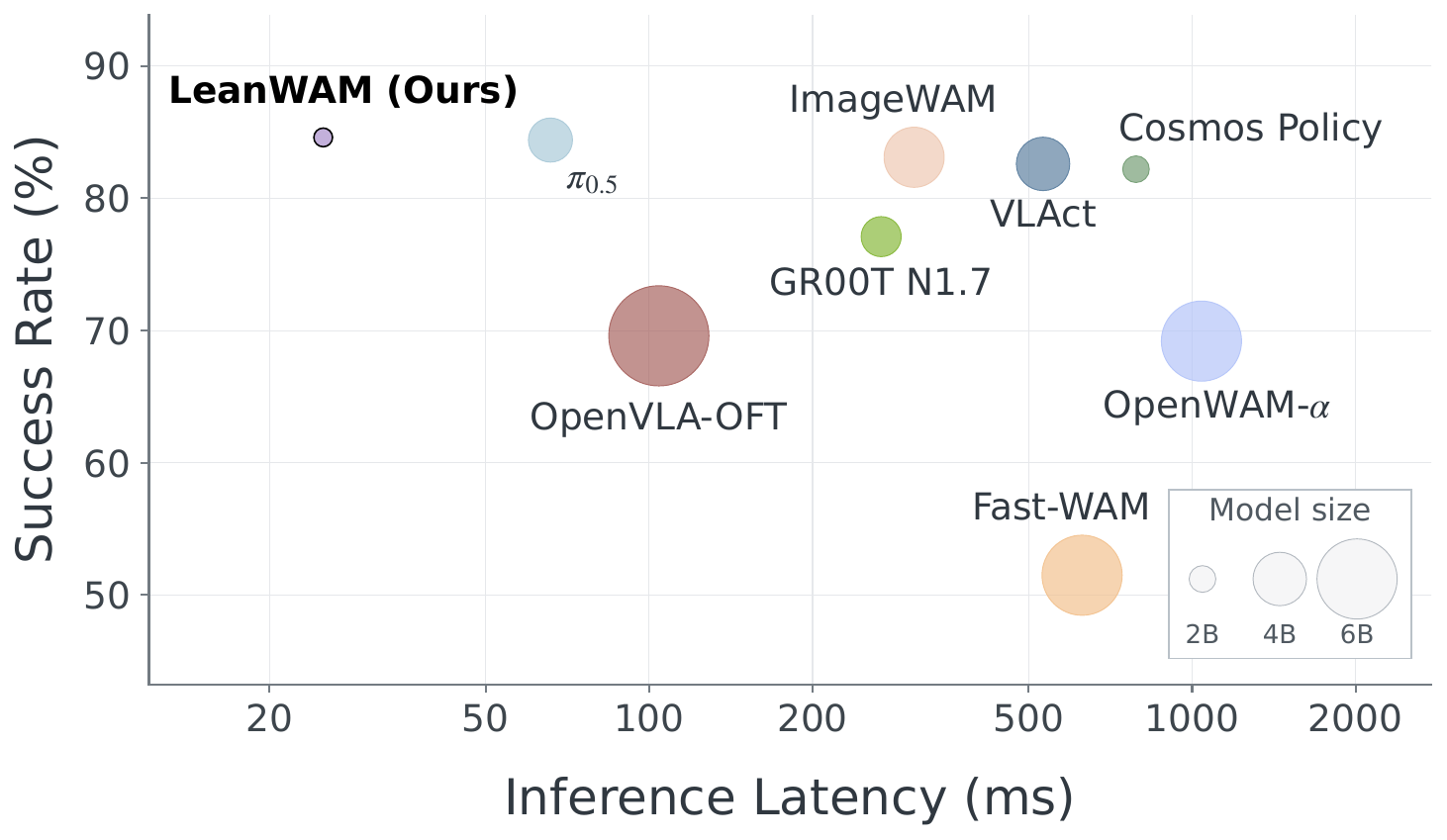}
\caption{\textbf{Success rate versus inference latency on LIBERO-Plus.} Latency per action chunk is shown on a log scale, and bubble size denotes model size.}
\label{fig:efficiency}
\end{figure}

\begin{figure*}[t]
\centering
\includegraphics[width=\textwidth]{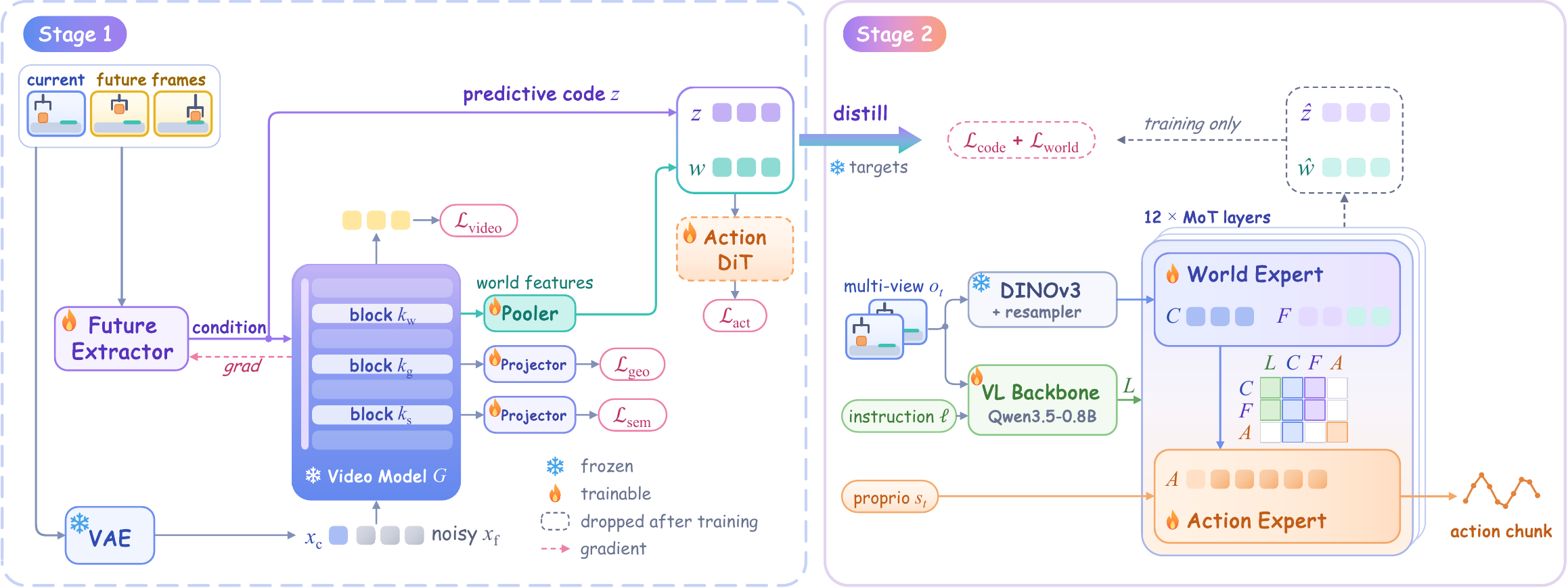}
\caption{\textbf{Overview of LeanWAM.}
\textbf{Stage~1:} A Future Extractor produces a predictive code $z$ that conditions a frozen video model $G$. Semantic, geometric, RGB, and action losses jointly shape the code. An attentive pooler extracts world tokens $\mathbf{w}$ from the hidden states of $G$, and an action DiT reads $[P_a(z);\mathbf{w}]$.
\textbf{Stage~2:} The World Expert processes current tokens $\mathbf{C}$ and learnable future queries $\mathbf{F}$ to predict $z$ and $\mathbf{w}$, while the Action Expert generates actions. Only the Stage-2 policy is deployed.}
\label{fig:overview}
\end{figure*}

In this paper, we ask: \textit{what information about the future does a policy need, and how compactly can it be represented?} Recent studies suggest that useful signals lie both in future observations and in the intermediate features of video models~\cite{vpp,liang2025video,wen2024vidman,invisiblehand}. Our probing study (Sec.~\ref{sec:probe}) shows that joint video--action training makes robot actions substantially easier to decode from video features. Combining these adapted features with VAE latents and semantic features improves decoding further. In contrast, adding features from the original pretrained video model provides no gain.
These combined features can also be compressed into a few tokens: a single aggregated token retains most of the action-decoding performance achieved with 32 tokens. Intermediate frames further improve decoding within the same action horizon and token budget. Together, these findings motivate a compact, action-supervised representation that combines complementary cues across a short clip.

We present LeanWAM, a latent world action model that transfers the priors of a frozen pretrained video generator into a lightweight policy. Figure~\ref{fig:overview} illustrates how LeanWAM learns compact predictive representations and distills them into the deployed policy. In Stage~1, a Future Extractor compresses a short clip into a predictive code that conditions the frozen video model. The current frame supplies scene context, while the code describes the transition. Semantic, geometric, and RGB supervision at increasing depths guides this description without additional generation branches. An action DiT reads both the predictive code and a compact readout of the model's hidden states, called world tokens. Action gradients shape this predictive interface through the frozen model. In Stage~2, a lightweight Mixture-of-Transformers policy learns to predict both representations from current observations while generating actions. Only this policy is deployed, and the video generator is used exclusively during training.

Our contributions are threefold:
\begin{itemize}[leftmargin=*,itemsep=2pt,topsep=2pt]
\item We introduce a two-stage framework that transfers a frozen video generator's priors through compact predictive codes and world tokens into a lean deployed policy, without updating or deploying the video generator.
\item Our probes show that a few tokens retain most of the action information in complementary future representations (Sec.~\ref{sec:probe}), and that the learned code supports better action decoding than dense future representations. Behavior analysis further reveals greater cross-task consistency with action supervision (Sec.~\ref{sec:analysis}).
\item Without large-scale embodied pretraining, our 1.4B-parameter policy surpasses strong vision-language-action (VLA) and WAM baselines on LIBERO~\cite{libero}, LIBERO-Plus~\cite{liberoplus}, and RoboTwin~2.0~\cite{robotwin2}. It predicts an action chunk in 25.1\,ms on an A100 GPU, with 12.3--41.5$\times$ lower latency than representative WAMs (Figure~\ref{fig:efficiency}).
\end{itemize}

\section{Related Work}
\label{sec:related}

\paragraph{Vision-language-action models.}
Vision-language-action (VLA) models transfer the knowledge of pretrained vision-language models to robot control. RT-2~\cite{rt2} and OpenVLA~\cite{openvla} represent actions as discrete tokens, while $\pi_0$~\cite{pi0} and GR00T N1~\cite{gr00tn1} pair a vision-language backbone with a continuous action generator. Other work scales training across tasks and data sources~\cite{pi05}, robot embodiments~\cite{lingbotvla}, memory augmentation~\cite{robomme}, or aligned heterogeneous manipulation data~\cite{qwenrobotmanip}. LeanWAM retains a language-conditioned continuous-action policy and uses future observations as additional training supervision.

\begin{figure*}[t]
\centering
\includegraphics[width=\textwidth]{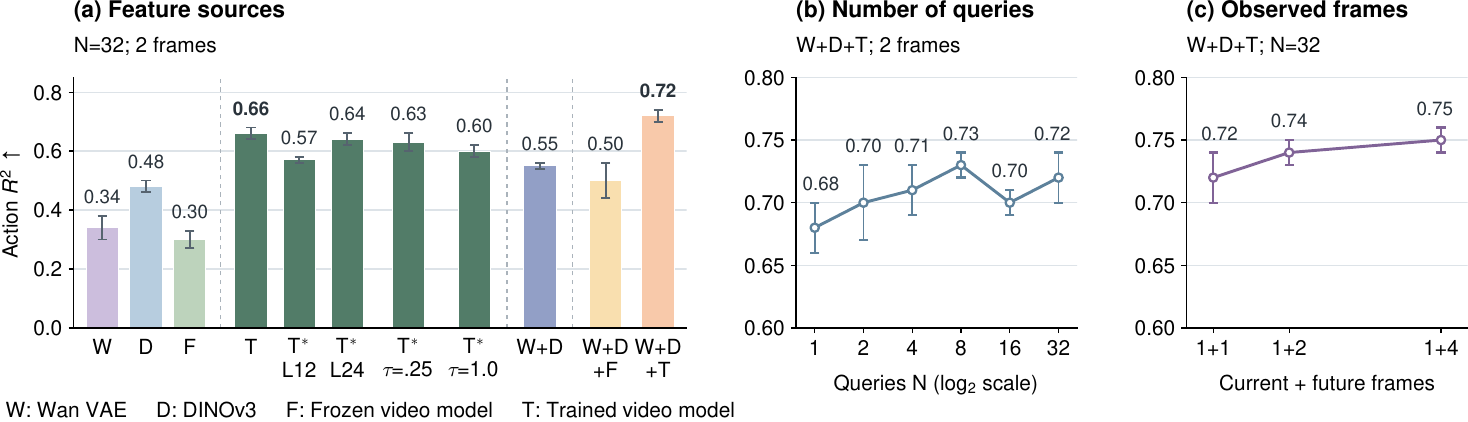}
\caption{\textbf{Probing what a policy needs from the future.} We report the action $R^2$ of the probe on held-out LIBERO episodes, averaged over three seeds with different data splits. Error bars show $\pm1$ standard deviation. \textbf{(a)} Feature sources. T$^*$ changes the default block 18 or noise level $\tau=0.75$. \textbf{(b)} Number of probe queries $N$. \textbf{(c)} Number of observed frames: the current frame plus 1, 2, or 4 future frames uniformly sampled over the same action horizon.}
\label{fig:probe}
\end{figure*}

\paragraph{World action models.}
WAMs couple action learning with the prediction of how the visual world evolves. DreamZero~\cite{dreamzero} adapts a 14B video generation model to jointly predict videos and actions. Motus~\cite{motus} combines understanding, video, and action experts in a Mixture-of-Transformers~\cite{mot}, and the LingBot-VA series~\cite{lingbotva,lingbotva2} studies causal modeling and native video--action pretraining. Flex-$\pi$~\cite{flexpi} and DreamWAM~\cite{dreamwam} extend future prediction to geometry, motion, or semantics. Fast-WAM~\cite{fastwam} retains video co-training while skipping future generation at inference. These approaches retain a video backbone at deployment. InternVLA-A1.5~\cite{internvlaa15} and PDRVLA~\cite{pdrvla} instead use frozen video models only during training. Recent analyses further show that, beyond the predicted future, intermediate video features also provide useful action representations~\cite{vpp,liang2025video,wen2024vidman,invisiblehand}.
LeanWAM learns a compact code and an action-supervised readout of these features, then distills both into a separate policy.

\paragraph{Latent world action models.}
WoG~\cite{wog} and Being-H0.7~\cite{beingh07} use future observations to supervise compact latents for action generation. LaWAM~\cite{lawam} predicts visual subgoals in a frozen feature space. VLA-JEPA~\cite{vlajepa} and JEPA-WAM~\cite{jepawam} learn policies with future-state prediction in pretrained video representation spaces. ReWAM~\cite{rewam} compresses DINOv3 features through a bottleneck trained only by action gradients, and GE-Act~2.0~\cite{geact2} aligns its visual latents with multiple perception teachers. LeanWAM complements these approaches by shaping its predictive code through a pretrained video generator, jointly using action, RGB, semantic, and geometric supervision.

\section{What Does a Policy Need from the Future?}
\label{sec:probe}

LeanWAM rests on the premise that a policy needs only a small, action-relevant part of the future. To test this premise, we train a probe to recover actions from representations of observed transitions. We evaluate action $R^2$, which measures the relative reduction in squared prediction error compared with predicting the mean action. Higher scores indicate that actions are easier to decode from a representation under this probe. We ask three questions: (1) Which future representations make actions easy to decode? (2) How many tokens are needed for effective action decoding? (3) Do intermediate frames provide useful temporal context?

\paragraph{Setup.}
We train an attentive probe~\cite{vjepa} to aggregate frozen features into $N$ tokens with learnable queries and decode an $H$-step action chunk with a small transformer. We compare Wan VAE~\cite{wan2025} latents (W), DINOv3~\cite{dinov3} patch features (D), and Cosmos-Predict2.5-2B~\cite{cosmos25} hidden states, before (F) and after (T) video--action training on the probe's training episodes. Each probe is trained separately with mean squared error on LIBERO~\cite{libero} demonstrations, using the same episode split and training schedule within each seed. By default, we use $N=32$ and the current and final frames of the chunk. For simplicity, we extract video features from a single transformer block. We use block 18 at noise level $\tau=0.75$, where $\tau=1$ denotes pure noise. Full implementation details are provided in Appendix~\ref{sec:supp_probe}.

\paragraph{Feature sources.}
Figure~\ref{fig:probe}(a) shows that DINOv3 features yield higher
action $R^2$ than VAE latents, consistent with broader evidence that semantic latents support action recovery~\cite{reconsem}. The pretrained video features F have the lowest mean among the tested single sources ($R^2=0.30$), whereas features of the same model after training on robot data (T) reach $0.66$. This gap shows that joint video--action training on robot demonstrations makes the model's features more accessible for action decoding. The sources also provide complementary cues: W+D reaches 0.55, above either W or D alone, and adding T raises $R^2$ to 0.72, above both W+D and T. Adding F instead gives a lower mean of 0.50, indicating that more feature sources do not necessarily improve decoding.

Where the video features are read also matters. Block 18 at $\tau=0.75$ has the highest mean $R^2$ among the tested settings. Moving to block 12 reduces the score to 0.57, while block 24 remains close at 0.64. At block 18, both a lower noise level ($\tau=0.25$) and pure noise ($\tau=1$) reduce decoding performance. These results indicate that action decodability depends on both depth and noise level, consistent with prior analyses of diffusion representations~\cite{ddae,gen4u}.

\paragraph{Number of queries.}
Figure~\ref{fig:probe}(b) varies the number of queries $N$ for W+D+T. Each query is a learnable vector with the probe's hidden width of 512, and the queries jointly aggregate clip features through cross-attention. Although the probe receives more than a thousand feature tokens, a single summary token already yields $R^2=0.68$, close to the $0.72$ with 32 tokens. As $N$ grows, $R^2$ stays within a narrow range and peaks at $N=8$, with no consistent gain beyond a few queries. These results support learning a compact representation of observed transitions that retains most of the action-decoding performance.

\paragraph{Temporal sampling.}
Figure~\ref{fig:probe}(c) compares two, three, and five frames for W+D+T over the same action horizon, with $N=32$ throughout. All settings include the current frame and the final frame. The three-frame setting adds the midpoint, and the five-frame setting samples at quarter intervals. The corresponding mean $R^2$ scores are 0.72, 0.74, and 0.75. Denser observations therefore improve decoding without extending the horizon or increasing the number of queries. The endpoints capture the overall visual change but leave the intervening motion ambiguous, and the gains are consistent with intermediate frames supplying such motion cues.

\paragraph{Implications.}
These results suggest that a small set of tokens can retain complementary action cues from dense future representations, while intermediate frames provide additional temporal context. They also expose a challenge: joint video--action training improves video features for action decoding, but updating a large video model is costly. LeanWAM instead learns a compact predictive code shaped by visual and action supervision through a frozen video generator. We evaluate the resulting representations and closed-loop policies in Sec.~\ref{sec:exp}.

\section{Method}
\label{sec:method}

At time $t$, a policy receives camera observations $o_t$, a proprioceptive state $s_t$, and a language instruction $\ell$, and predicts an action chunk $\mathbf{a}_t=(a_t,\ldots,a_{t+H-1})$. Motivated by Sec.~\ref{sec:probe}, training also uses a clip $\mathbf{V}_t$ containing the current frame and several future frames uniformly sampled through $t+H$. LeanWAM is trained in two stages (Figure~\ref{fig:overview}). Stage~1 encodes the observed transition into a predictive code and world tokens (Sec.~\ref{sec:stage1}). Stage~2 trains a lean policy to predict the code and world tokens from the current observation (Sec.~\ref{sec:stage2}), and only this policy is deployed.

Video and action objectives use flow matching~\cite{lipman2023flow,liu2023rectified}. Given a clean sample $\mathbf{x}$, Gaussian noise $\boldsymbol{\epsilon}\sim\mathcal{N}(\mathbf{0},\mathbf{I})$, and a flow time $\tau\in[0,1]$, let $\mathbf{x}^{\tau}=(1-\tau)\,\mathbf{x}+\tau\,\boldsymbol{\epsilon}$. A network $f$ with condition $\mathbf{c}$ regresses the velocity $\boldsymbol{\epsilon}-\mathbf{x}$:
\begin{equation}
\mathcal{L}_{\mathrm{FM}}(f;\mathbf{x},\mathbf{c})=\mathbb{E}_{\tau,\boldsymbol{\epsilon}}\Big[\big\|f(\mathbf{x}^{\tau},\tau,\mathbf{c})-(\boldsymbol{\epsilon}-\mathbf{x})\big\|_2^2\Big],
\label{eq:fm}
\end{equation}
and generates samples by integrating from $\tau=1$ to $\tau=0$.

\subsection{Learning Predictive Codes from a Frozen Video Model}
\label{sec:stage1}

We use Cosmos-Predict2.5-2B~\cite{cosmos25} as the frozen video model $G$. Its frozen Wan VAE encoder compresses time by $4\times$, producing current and future latents $\mathbf{x}_{\mathrm{c}}$ and $\mathbf{x}_{\mathrm{f}}$.

\paragraph{Encoding the transition.}
The Future Extractor $E_\theta$ maps the clip to a predictive code $z=E_\theta(\mathbf{V}_t)\in\mathbb{R}^{N_z\times d_z}$ through two frozen encoders. The VAE latents lie in the latent space of $G$ and preserve fine-grained appearance and motion, while DINOv3~\cite{dinov3} provides semantic patch features that are more robust to visual distribution shifts. The two are complementary for action decoding (Sec.~\ref{sec:probe}). Tokens of both encoders are mapped to a common width and spatial grid, and fixed spatiotemporal sinusoidal position embeddings are added. An 18-layer Q-Former~\cite{blip2} aggregates these tokens through self-attention and cross-attention with $N_z$ learnable queries. A linear layer projects each query to $d_z$ dimensions. We set $N_z=d_z=32$, encoding the transition over the action horizon in $1{,}024$ scalar values.

\paragraph{Explaining the future to a frozen video model.}
We replace the text context of $G$ with $P_g(z)$ in every cross-attention block, where $P_g$ is a trainable linear projection into the text-context space of $G$. Conditioned on the clean current latent, $G$ predicts the future through
\begin{equation}
\mathcal{L}_{\mathrm{video}}=\mathcal{L}_{\mathrm{FM}}\big(G;\ \mathbf{x}_{\mathrm{f}},\ (\mathbf{x}_{\mathrm{c}},P_g(z))\big).
\label{eq:video}
\end{equation}
The weights of $G$ remain fixed, but gradients pass through $G$ to train the Future Extractor. Since a frozen $G$ interprets its condition in a fixed way, the code can lower the loss only by providing additional transition information that helps the frozen model predict the future. The current frame supplies scene appearance, and the pretrained generator supplies visual priors. This encourages the limited code capacity to describe changes needed to explain the observed future. The code thus learns to guide an existing generative model rather than co-adapt with it, grounding representation learning in pretrained spatiotemporal priors that distillation later transfers to the policy.

\paragraph{Semantic and geometric alignment.}
To complement RGB prediction, we align intermediate features of $G$ with frozen perception teachers~\cite{repa}. Block $k_{\mathrm{s}}=9$ is aligned with DINOv3 patch features, and block $k_{\mathrm{g}}=14$ is aligned with patch features from the global-attention blocks of VGGT-$\Omega$~\cite{vggtomega}. Both teachers encode the clean clip. We keep the current-frame features separate and linearly interpolate future features in time to match $G$'s token layout. A trainable projector $P_{\mathrm{sem}}$, comprising a 3D convolution and an MLP, maps hidden states $\mathbf{h}^{(k_{\mathrm{s}})}$ to the semantic target space, and we maximize their patch-wise cosine similarity:
\begin{equation}
\mathcal{L}_{\mathrm{sem}}=\frac{1}{|\Omega|}\sum_{i\in\Omega}\Big[1-\cos\big(P_{\mathrm{sem}}(\mathbf{h}^{(k_{\mathrm{s}})})_i,\ \widetilde{\mathbf{Y}}^{\mathrm{sem}}_i\big)\Big],
\label{eq:sem}
\end{equation}
where $\Omega$ indexes spatiotemporal tokens and $\widetilde{\mathbf{Y}}^{\mathrm{sem}}$ denotes the resampled targets. The geometric loss $\mathcal{L}_{\mathrm{geo}}$ uses an analogous projector $P_{\mathrm{geo}}$ and VGGT-$\Omega$ targets $\widetilde{\mathbf{Y}}^{\mathrm{geo}}$.

The supervision follows the depth of $G$: semantics at an early block, geometry at a middle block that builds on the semantic layout, and RGB at the output. Since video generation models already encode much of the semantics, geometry, and motion of a scene~\cite{gen4u}, the code does not need to store high-dimensional feature maps with their backgrounds. It only needs to steer $G$ toward internal states consistent with the true future, which also avoids the conflicts that may arise when a small latent is regressed onto several teachers directly. All three losses share one forward pass of $G$, without separately generating semantic or geometric outputs.

\paragraph{Action supervision through the frozen video model.}
An action DiT $\pi_\psi$ reads the code. Its inputs are proprioception and noisy action tokens, with flow time supplied through adaptive layer normalization (adaLN)~\cite{dit}. Self-attention alternates with cross-attention to the conditioning tokens. Besides the code, the action DiT receives a compact readout of $G$. During denoising, video-model hidden states encode physical plausibility~\cite{invisiblehand}, but predictive features do not necessarily make actions easy to decode.
Our probe shows that joint video--action training improves action decodability (Sec.~\ref{sec:probe}), motivating action-guided feature learning.
We achieve this without updating $G$: action supervision trains the code and its readout, shaping the code-conditioned features of $G$ while keeping its weights frozen. From the same forward pass of $G$, we take the future-position hidden states $\mathbf{h}^{(k_{\mathrm{w}})}_{\mathrm{f}}$ of block $k_{\mathrm{w}}=18$, which we call world features. An attentive pooler $R_\eta$ with $N_w=32$ learnable queries and 3 cross-attention layers compresses them into world tokens $\mathbf{w}=R_\eta(\mathbf{h}^{(k_{\mathrm{w}})}_{\mathrm{f}})\in\mathbb{R}^{N_w\times d_w}$, which are extracted downstream of both alignment blocks. The action objective is:
\begin{equation}
\mathcal{L}_{\mathrm{act}}^{(1)}
=
\mathcal{L}_{\mathrm{FM}}
\big(\pi_\psi;\ \mathbf{a}_t,\
(s_t,[P_a(z);\,\mathbf{w}])\big),
\label{eq:act1}
\end{equation}
where $P_a$ is a trainable linear projection from $d_z$ to $d_w$ dimensions, $[\cdot;\cdot]$ concatenates tokens, and $\mathbf{w}$ uses the noise level sampled for the video loss. The action gradients reach the code both directly and through $G$. This encourages controllable changes to remain accessible, since visual prediction alone can favor nuisance variation~\cite{whatdolam}. The action DiT is discarded after Stage~1.

\paragraph{Objective.}
Stage~1 minimizes
\begin{equation}
\mathcal{L}_{\mathrm{S1}}
=
\mathcal{L}_{\mathrm{act}}^{(1)}
+
\mathcal{L}_{\mathrm{video}}
+
\lambda_{\mathrm{sem}}\mathcal{L}_{\mathrm{sem}}
+
\lambda_{\mathrm{geo}}\mathcal{L}_{\mathrm{geo}},
\label{eq:stage1}
\end{equation}
with $\lambda_{\mathrm{sem}}=\lambda_{\mathrm{geo}}=0.1$. As in conventional WAMs, action and video supervision are combined with a pretrained video prior, but here they shape a compact predictive code while the video model remains frozen.

\subsection{Distilling the Future into a Lean Policy}
\label{sec:stage2}

Stage~1 encodes future observations available only during training. Stage~2 learns to predict these representations from current observations.

\paragraph{Architecture.}
The policy consists of a vision-language backbone, a World Expert, and an Action Expert. The backbone, Qwen3.5-0.8B~\cite{qwen35} truncated to the first 12 of its 24 layers, encodes the multi-view observation and the instruction into context $\mathbf{L}$. The two experts are 12-layer transformers coupled as a Mixture-of-Transformers~\cite{mot}, with separate weights and masked joint self-attention at every layer. Only the World Expert can attend to the backbone context $\mathbf{L}$, which is supplied through layer-specific normalization and key/value projections. The World Expert holds two groups of tokens. The current tokens $\mathbf{C}$ describe the present scene: each camera view is encoded by a frozen DINOv3 and tagged with a learnable view embedding, and a visual resampler compresses all views into 64 tokens. The future queries $\mathbf{F}$ are $N_z+N_w$ learnable tokens that predict the code and the world tokens. The Action Expert processes the proprioception and noisy action tokens $\mathbf{A}$, conditioned on flow time through adaLN. Detailed module architectures for both training stages are provided in Appendix~\ref{sec:supp_arch}.

\paragraph{Attention mask.}
The backbone keeps its causal mask and does not attend to the experts. Both $\mathbf{C}$ and $\mathbf{F}$ attend to $\{\mathbf{L},\mathbf{C},\mathbf{F}\}$, allowing the future queries to predict from the observation and instruction while refining the current tokens with predictive information. The Action Expert attends only to $\{\mathbf{C},\mathbf{A}\}$, with no direct access to $\mathbf{L}$ or $\mathbf{F}$. Language and predicted-future information therefore influence actions through $\mathbf{C}$, whose computation is independent of action denoising.

\paragraph{Distillation.}
At the last layer, separate layer-normalization and linear heads map the future queries to code predictions $\hat{z}$ and world-token predictions $\hat{\mathbf{w}}$. The targets come from the frozen Stage-1 modules: $\bar{z}=\mathrm{LN}(E_\theta(\mathbf{V}_t))$ and $\bar{\mathbf{w}}=\mathrm{LN}(R_\eta(\mathbf{h}^{(k_{\mathrm{w}})}_{\mathrm{f}}))$, where $\mathrm{LN}$ is a token-wise layer normalization without affine parameters. To obtain $\bar{\mathbf{w}}$, $G$ receives $P_g(z)$, the clean current latent, and the future latent corrupted at $\tau^\star=0.75$, selected from our probe. We use the same noise level for target extraction across samples. We minimize $\mathcal{L}_{\mathrm{code}}=\mathrm{SmoothL1}(\hat{z},\bar{z})$ and $\mathcal{L}_{\mathrm{world}}=\mathrm{SmoothL1}(\hat{\mathbf{w}},\bar{\mathbf{w}})$. The two distillation targets are complementary: the code is jointly shaped by all Stage-1 objectives, while the world tokens retain spatial structure from the video prior.

\paragraph{Objective.}
Let $\mathcal{L}_{\mathrm{act}}^{(2)}$ denote the flow-matching action objective of the Stage-2 policy. We randomly initialize the Action Expert and freeze all
Stage-1 modules. Stage~2 minimizes
\begin{equation}
\mathcal{L}_{\mathrm{S2}}
=
\mathcal{L}_{\mathrm{act}}^{(2)}
+
\lambda_{\mathrm{code}}\mathcal{L}_{\mathrm{code}}
+
\lambda_{\mathrm{world}}\mathcal{L}_{\mathrm{world}},
\label{eq:stage2}
\end{equation}
where $\lambda_{\mathrm{code}}=\lambda_{\mathrm{world}}=0.1$.


\paragraph{Inference.}
The backbone and World Expert each run once per observation. We cache the layer-wise keys and values of $\mathbf{C}$ and run only the Action Expert for ten denoising steps. All Stage-1 modules and the distillation heads are discarded. The deployed policy has 1.4B parameters and renders no future video.

\section{Experiments}
\label{sec:exp}

We evaluate LeanWAM on LIBERO~\cite{libero}, LIBERO-Plus~\cite{liberoplus}, and RoboTwin~2.0~\cite{robotwin2}, comparing success rates and latency with VLA and WAM baselines (Sec.~\ref{sec:main_results}), analyzing predictive representations (Sec.~\ref{sec:analysis}), and ablating model components (Sec.~\ref{sec:ablation}).

\subsection{Experimental Setup}
\label{sec:exp_setup}

\paragraph{Benchmarks.}
We use four 10-task LIBERO suites~\cite{libero}: Spatial, Object, Goal, and Long. A single model is trained jointly on these suites and evaluated on 500 episodes per suite. LIBERO-Plus~\cite{liberoplus} adds seven perturbation types: camera viewpoint, robot initial state, language, lighting, background, sensor noise, and object layout. We evaluate the LIBERO-trained checkpoint on all 10,030 episodes without further training. RoboTwin~2.0~\cite{robotwin2} contains 50 bimanual tasks with clean and randomized settings, and randomization varies clutter, background, lighting, table height, and instructions. We jointly train on clean and randomized demonstrations of 10 selected tasks and evaluate 100 episodes per task per setting.

\paragraph{Implementation details.}
We train both stages for 30 epochs on LIBERO, and Stage~1 and Stage~2 for 10 and 15 epochs on RoboTwin~2.0. We report success rates (\%). Baseline success rates are taken from the corresponding papers or public benchmark reports. We separately measure inference latency for representative VLA and WAM baselines. Further details are in
Appendices~\ref{sec:supp_exp} and~\ref{sec:supp_efficiency}.

\subsection{Main Results}
\label{sec:main_results}
 
\begin{table}[t]
\centering
\footnotesize
\setlength{\tabcolsep}{3.1pt}
\caption{\textbf{Results on LIBERO.} Success rates (\%) over 500 episodes per suite. Size: number of parameters. PT: large-scale embodied pretraining. \textbf{Bold}: best. \underline{Underline}: second best.}
\label{tab:libero}
\begin{tabular}{lccccccc}
\toprule
Method & Size & PT & Spatial & Object & Goal & Long & Avg. \\
\midrule
\multicolumn{8}{l}{\textit{Vision-language-action models}} \\
OpenVLA-OFT~\cite{openvlaoft} & 7.5B & \cmark & 97.6 & 98.4 & \underline{97.9} & 94.5 & 97.1 \\
$\pi_{0.5}$~\cite{pi05} & 3.3B & \cmark & \textbf{98.8} & 98.2 & \textbf{98.0} & 92.4 & 96.9 \\
GR00T N1.7~\cite{gr00tn1} & 3B & \cmark & 97.7 & 98.5 & 97.5 & 94.4 & 97.0 \\
LAPA~\cite{lapa} & 7B & \cmark & 73.8 & 74.6 & 58.8 & 55.4 & 65.7 \\
UniVLA~\cite{univla} & 7B & \cmark & 96.5 & 96.8 & 95.6 & 92.0 & 95.2 \\
VLA-JEPA~\cite{vlajepa} & 2.3B & \cmark & 96.2 & 99.6 & 97.2 & 95.8 & 97.2 \\
StarVLA-$\pi$~\cite{starvla} & 7.8B & \xmark & \textbf{98.8} & 99.6 & 95.8 & 88.4 & 95.7 \\
\midrule
\multicolumn{8}{l}{\textit{World action models}} \\
Fast-WAM~\cite{fastwam} & 6B & \xmark & 98.2 & \textbf{100.0} & 97.0 & 95.2 & 97.6 \\
JEPA-WAM~\cite{jepawam} & 1.3B & \xmark & 95.6 & 99.4 & 97.2 & 94.6 & 96.7 \\
Motus~\cite{motus} & 8B & \cmark & 96.8 & \underline{99.8} & 96.6 & \underline{97.6} & \underline{97.7} \\
\midrule
\rowcolor{oursrow} \textbf{LeanWAM} (ours) & 1.4B & \xmark & \underline{98.4} & 99.4 & 97.8 & \textbf{98.0} & \textbf{98.4} \\
\bottomrule
\end{tabular}
\end{table}
 
\begin{table*}[t]
\centering
\footnotesize
\setlength{\tabcolsep}{6.6pt}
\caption{\textbf{Zero-shot generalization and latency on LIBERO-Plus.} Success rates (\%) by perturbation and overall (Total). LeanWAM is trained only on LIBERO. Latency per action chunk is measured for LeanWAM and representative VLA and WAM baselines on the same A100 GPU at batch size 1; -- indicates that latency was not measured. Other notation follows Table~\ref{tab:libero}.}
\label{tab:liberoplus}
\begin{tabular}{lccccccccccc}
\toprule
Method & Size & PT & Latency (ms) & Camera & Robot & Language & Light & Background & Noise & Layout & Total \\
\midrule
\multicolumn{12}{l}{\textit{Vision-language-action models}} \\
OpenVLA-OFT~\cite{openvlaoft} & 7.5B & \cmark & 104.2 & 56.4 & 31.9 & 79.5 & 88.7 & 93.3 & 75.8 & 74.2 & 69.6 \\
$\pi_0$~\cite{pi0} & 3.3B & \cmark & -- & 13.8 & 6.0 & 58.8 & 85.0 & 81.4 & 79.0 & 68.9 & 53.6 \\
$\pi_0$-FAST~\cite{pi0fast} & 3B & \cmark & -- & 65.1 & 21.6 & 61.0 & 73.2 & 73.2 & 74.4 & 68.8 & 61.6 \\
$\pi_{0.5}$~\cite{pi05} & 3.3B & \cmark & \underline{65.8} & 78.4 & 73.6 & 80.8 & 96.2 & 94.1 & 89.0 & 84.5 & \underline{84.4} \\
VLA-JEPA~\cite{vlajepa} & 2.3B & \cmark & -- & 63.3 & 67.1 & 85.4 & 95.6 & 93.6 & 66.3 & 85.1 & 79.5 \\
StarVLA~\cite{starvla} & 4B & \xmark & -- & 52.5 & 49.8 & 88.5 & 95.7 & \underline{95.7} & 73.0 & 76.9 & 74.1 \\
VLAct~\cite{vlact} & 4B & \cmark & 531.4 & 73.9 & 68.4 & 81.5 & 96.7 & \textbf{96.7} & 86.0 & 83.3 & 82.6 \\
\midrule
\multicolumn{12}{l}{\textit{World action models}} \\
Fast-WAM~\cite{fastwam} & 6B & \xmark & 627.1 & 16.4 & 44.5 & 68.9 & 78.2 & 53.7 & 37.7 & 60.7 & 51.5 \\
OpenWAM-$\alpha$~\cite{openwam} & 6B & \cmark & 1040.3 & 33.8 & \underline{76.1} & 88.0 & 97.0 & 87.1 & 39.8 & 77.5 & 69.2 \\
4D-WAM~\cite{fourwam} & 6B & \xmark & -- & 45.2 & 64.3 & 90.6 & 94.3 & 57.7 & 69.1 & 79.2 & 71.0 \\
ST-WAM~\cite{stwam} & 7B & \xmark & -- & 55.4 & 60.1 & 79.3 & 93.0 & 74.2 & 79.5 & 74.3 & 72.8 \\
Faster-WAM~\cite{fasterwam} & 5B & \xmark & -- & 67.9 & 49.0 & \textbf{92.1} & 94.3 & 57.0 & 82.3 & 82.7 & 75.0 \\
Cosmos Policy~\cite{cosmospolicy} & 2B & \xmark & 788.2 & 75.8 & 63.3 & 81.7 & 96.5 & 88.9 & 92.7 & 82.2 & 82.2 \\
JEPA-WAM~\cite{jepawam} & 1.3B & \xmark & -- & 79.2 & 59.2 & 68.2 & 93.3 & 94.6 & 83.6 & 76.1 & 79.2 \\
ImageWAM~\cite{imagewam} & 4.5B & \xmark & 307.6 & 80.8 & 50.3 & \underline{91.4} & \textbf{98.1} & 85.5 & \underline{93.8} & 80.5 & 83.1 \\
ABot-M0.5~\cite{abotm05} & 5B & \cmark & -- & 70.5 & \textbf{87.4} & 88.6 & 94.0 & 89.7 & 75.5 & \underline{85.2} & 83.4 \\
Being-H0.7~\cite{beingh07} & 3B & \cmark & -- & \underline{82.0} & 59.0 & 82.8 & \underline{97.8} & 90.0 & 93.5 & \textbf{88.5} & 82.1 \\
\midrule
\rowcolor{oursrow} \textbf{LeanWAM} (ours) & 1.4B & \xmark & \textbf{25.1} & \textbf{90.9} & 61.8 & 81.3 & 96.4 & 88.4 & \textbf{97.1} & 79.8 & \textbf{84.6} \\
\bottomrule
\end{tabular}
\end{table*}

\begin{table}[t]
\centering
\footnotesize
\setlength{\tabcolsep}{9pt}
\caption{\textbf{Results on RoboTwin 2.0.} Success rates (\%) averaged over 10 bimanual tasks, with 100 episodes per task and setting.}
\label{tab:robotwin}
\begin{tabular}{lccc}
\toprule
Method & Clean & Randomized & Avg. \\
\midrule
X-VLA~\cite{xvla} & 76.4 & 73.6 & 75.0 \\
$\pi_{0.5}$~\cite{pi05} & 83.5 & 80.3 & 81.9 \\
\midrule
Motus~\cite{motus} & 89.5 & 89.4 & \underline{89.5} \\
Fast-WAM~\cite{fastwam} & 89.3 & \underline{89.6} & \underline{89.5} \\
GigaWorld-Policy~\cite{gigaworldpolicy} & 89.2 & 87.8 & 88.5 \\
LingBot-VA~\cite{lingbotva} & \underline{89.9} & 87.4 & 88.7 \\
\midrule
\rowcolor{oursrow} \textbf{LeanWAM} (ours) & \textbf{93.9} & \textbf{92.3} & \textbf{93.1} \\
\bottomrule
\end{tabular}
\end{table}
 
\paragraph{LIBERO.}
LeanWAM reaches 98.4\% average success, the highest in Table~\ref{tab:libero}, without large-scale embodied pretraining. It also leads on Long at 98.0\%, 5.6 points above $\pi_{0.5}$, and exceeds Fast-WAM and Motus with roughly one-quarter to one-sixth as many parameters. The following benchmarks test robustness beyond these high in-distribution scores.

\paragraph{LIBERO-Plus.}
LIBERO-Plus tests generalization to perturbations unseen in training (Table~\ref{tab:liberoplus}). Without large-scale embodied pretraining, LeanWAM achieves the highest total success rate, outperforming all other WAMs in the table and $\pi_{0.5}$. It is particularly robust to camera and noise perturbations, leading the next best methods by 8.9 and 3.3 points. Several WAMs built on large video models are fragile under such perturbations, with Fast-WAM reaching only 16.4\% under camera changes. We hypothesize that LeanWAM is robust because its predictive code emphasizes manipulation-relevant regions and behaviors (Sec.~\ref{sec:analysis}) rather than scene appearance. Perturbed initial robot states remain challenging for LeanWAM, and the five strongest methods on this dimension all use large-scale embodied pretraining, suggesting that broader pretraining could complement our approach.

\paragraph{Efficiency.}
Table~\ref{tab:liberoplus} also reports inference latency. We measure latency for representative VLA and WAM baselines. LeanWAM achieves the lowest latency in this comparison at 25.1\,ms per action chunk. It is 4.2$\times$ faster than OpenVLA-OFT, which decodes a chunk in a single pass, 2.6$\times$ faster than $\pi_{0.5}$, and 12.3$\times$ to 41.5$\times$ faster than the four representative WAMs. Unlike these WAMs, LeanWAM does not run a generative visual backbone at inference: its backbone and World Expert run once per observation, and only the Action Expert iterates over ten denoising steps. Measurement details and our inference implementation are in Appendix~\ref{sec:supp_efficiency}.

\paragraph{RoboTwin 2.0.}
On RoboTwin~2.0 (Table~\ref{tab:robotwin}), LeanWAM leads both clean and randomized evaluations on our 10-task subset. It improves the average by 3.6 points over the strongest baselines, Motus and Fast-WAM, which run billion-parameter video models. Per-task results are in Appendix~\ref{sec:supp_robotwin}.

\begin{table}[t]
\centering
\small
\setlength{\tabcolsep}{6pt}
\caption{\textbf{Probing LeanWAM's representations.} Action $R^2$ for predictive code (P), world tokens (V), and their combination, compared with the best dense combination in Sec.~\ref{sec:probe}. Probes follow Sec.~\ref{sec:probe}, with results averaged over three seeds. $\Delta$: change relative to W+D+T.}
\label{tab:probe_learned}
\begin{tabular}{lccc}
\toprule
Representation & Tokens & Action $R^2$ & $\Delta$ \\
\midrule
W+D+T (dense, Sec.~\ref{sec:probe}) & $>$3k & 0.755 & -- \\
\midrule
P (predictive code) & 32 & \underline{0.805} & {\color{ForestGreen}$+$0.050} \\
V (world tokens) & 32 & 0.622 & {\color{BrickRed}$-$0.133} \\
P+V & 64 & \textbf{0.843} & {\color{ForestGreen}$+$0.088} \\
\bottomrule
\end{tabular}

\end{table}

\subsection{What Does the Predictive Code Capture?}
\label{sec:analysis}

\paragraph{Probing the learned representations.}
We rerun Stage~1 using only the probe's 400 training episodes, extract predictive codes (P) and world tokens (V) from five-frame clips, and train separate probes on P, V, and P+V. Evaluation uses the held-out episodes and protocol of Sec.~\ref{sec:probe}.
 
Table~\ref{tab:probe_learned} shows that the 32-token code reaches $R^2=0.805$, above the dense W+D+T combination at 0.755. Stage~1 therefore makes action information more accessible through a compact representation. World tokens score lower alone (0.622), yet their $R^2$ is more than twice that of F ($0.30$), read from the same video model under text conditioning, and close to that of T ($0.66$), read after training the model. Although V uses five frames whereas F and T use two, the observed gain over F is substantial. As with the dense sources in Sec.~\ref{sec:probe}, the two are complementary: with the same probe capacity, P+V performs best ($0.843$), suggesting that the world tokens carry action cues not readily decodable from the code. This supports distilling both targets into the policy. The corresponding closed-loop ablation appears in Sec.~\ref{sec:ablation}.

\begin{figure}[t]
\centering
\includegraphics[width=\columnwidth]{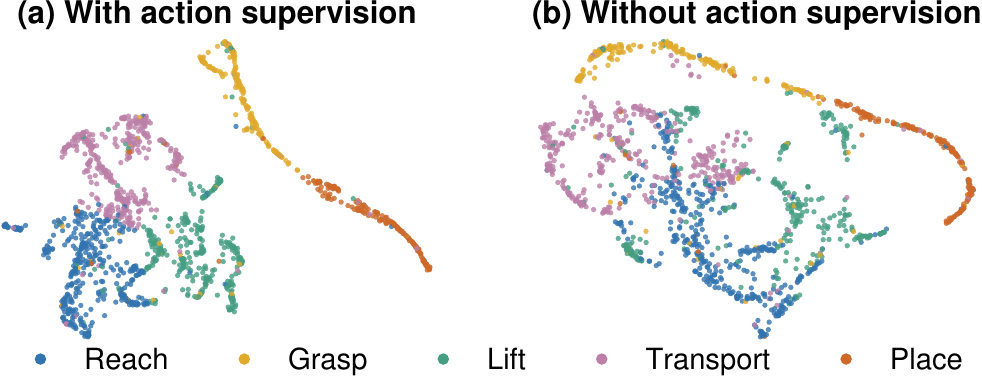}
\caption{\textbf{Predictive codes with and without action supervision.} The same 1,500 LIBERO clips, colored by behavior and projected with PCA followed by UMAP~\cite{umap}. \textbf{(a)} Full Stage-1 objective. \textbf{(b)} Without action loss. Reach, Lift, and Transport form more coherent neighborhoods in (a).}
\label{fig:umap}
\end{figure}

\begin{figure}[t]
\centering
\includegraphics[width=\columnwidth]{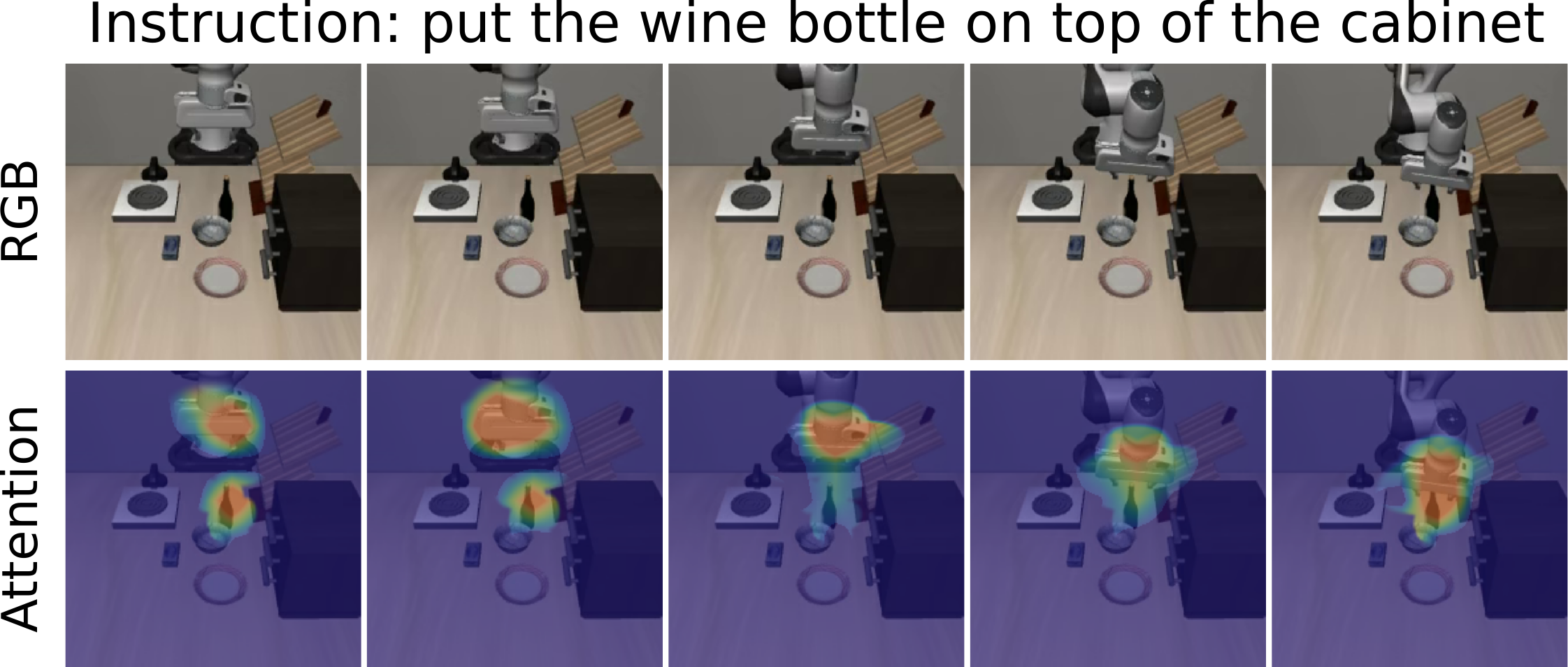}
\caption{\textbf{Query attention of the Future Extractor.} Top: demonstration frames. Bottom: query cross-attention to the DINOv3 tokens. The queries concentrate on the gripper and the target bottle.}
\label{fig:attn}
\end{figure}

\paragraph{Behavior structure of the code.}
We study how action supervision in Stage~1 shapes the code. We sample 1,500 non-overlapping clips balanced across LIBERO tasks and label each with one of five behaviors, Reach, Grasp, Lift, Transport, or Place, using Gemini~3.6 Flash~\cite{gemini36flash}. The same clips are encoded by Future Extractors trained with and without the Stage-1 action loss. Figure~\ref{fig:umap} visualizes the two sets of codes with UMAP~\cite{umap}. With action supervision, codes of the same behavior form distinct clusters or more coherent neighborhoods. Without it, Grasp and Place still lie apart, but Reach, Lift, and Transport intermix. Supervision through the video model alone thus captures salient events such as grasping and releasing, while the action loss improves separation among these behaviors.

Since UMAP can distort distances, we quantify this structure in the original 1,024-dimensional space. For each code, we find 15 cosine-nearest neighbors from other tasks, measure their behavior-label agreement, and macro-average over behaviors. Restricting neighbors to other tasks reduces task-specific similarity. Agreement rises from 70.7\% to 78.5\% with action supervision, versus an approximately 20\% random-neighbor reference. This supports more consistent behavior representations across tasks beyond the two-dimensional visualization.

\begin{table}[t]
\centering
\small
\setlength{\tabcolsep}{7pt}
\caption{\textbf{Ablations on LIBERO-Plus.} Total success rates (\%). $\Delta$ is relative to LeanWAM. $-$/$\rightarrow$: removal/replacement. $^\dagger$Also removes world tokens, leaving only code distillation in Stage~2.}
\label{tab:ablation}
\begin{tabular}{lcc}
\toprule
Variant & Total & $\Delta$ \\
\midrule
\rowcolor{oursrow} \textbf{LeanWAM} & \textbf{84.6} & -- \\
\midrule
\multicolumn{3}{l}{\textit{Stage 1}} \\
$-$ action supervision$^\dagger$ & 80.1 & $-$4.5 \\
$-$ video model$^\dagger$ & 81.0 & $-$3.6 \\
$-$ semantic and geometric alignment & 82.2 & $-$2.4 \\
frozen $\rightarrow$ trainable video model & 79.6 & $-$5.0 \\
alignment $\rightarrow$ prediction heads & 78.0 & $-$6.6 \\
\midrule
\multicolumn{3}{l}{\textit{Stage 2}} \\
$-$ code distillation & 79.3 & $-$5.3 \\
$-$ world distillation & 80.4 & $-$4.2 \\
$-$ code and world distillation & 75.8 & $-$8.8 \\
DINOv3 tokens $\rightarrow$ learnable tokens & 79.4 & $-$5.2 \\
\bottomrule
\end{tabular}
\end{table}

\paragraph{Where the Future Extractor attends.}
We visualize the cross-attention from the queries of the Future Extractor to the DINOv3 tokens of a LIBERO demonstration (Figure~\ref{fig:attn}). The attention concentrates on the gripper and the wine bottle that the robot is asked to move. The table, the background, and the other objects receive little attention. This example illustrates how the queries select manipulation-relevant regions when aggregating clip features. Additional attention visualizations are provided in Appendix~\ref{sec:supp_attn}.

\subsection{Ablation Studies}
\label{sec:ablation}
 
Table~\ref{tab:ablation} ablates the components of LeanWAM on all 10,030 LIBERO-Plus episodes. Each variant modifies a component or training pathway and otherwise follows Sec.~\ref{sec:exp_setup}, with details in Appendix~\ref{sec:supp_ablation}.

\paragraph{Stage 1.}
Removing action supervision or the video model reduces success by 4.5 and 3.6 points. These are coupled ablations: both also remove world tokens and retain only code distillation. They therefore test the corresponding training pathways, rather than isolating either component alone. Removing the semantic and geometric alignment costs 2.4 points. Replacing alignment with output heads that predict future DINOv3 features and depth, following WAMs that add semantic or geometric targets to the predicted future~\cite{flexpi,pdrvla}, is even worse (78.0\%). This supports our design: alignment only steers the internal states of the frozen model, whereas the heads must produce dense feature and depth maps from its last layer, which is specialized for RGB. Fine-tuning the video model, as most WAMs do, lowers the success rate to 79.6\%, although training makes video features more decodable in distribution (Sec.~\ref{sec:probe}). A trainable model can rely less on the code and may lose its general prior when fitted to limited robot data. Keeping it frozen is thus both cheaper and more robust.

\paragraph{Stage 2.}
With the same policy architecture and inference cost, action-only training reaches 75.8\%, 8.8 points below LeanWAM. Distilling world tokens alone or the code alone recovers part of this gap (79.3\% and 80.4\%), and jointly distilling them performs best, consistent with their complementarity in Sec.~\ref{sec:analysis}. Replacing current DINOv3 tokens with learnable tokens costs 5.2 points, supporting direct visual input to the World Expert alongside the vision-language backbone.
\section{Conclusion}
\label{sec:conclusion}

LeanWAM transfers video priors to control by learning compact predictive representations through a frozen video model. A lean policy predicts these representations from current observations, achieving strong task performance, robustness, and low inference latency without deploying the video model. 

\paragraph{Future work.} Our evaluation is limited to simulation. Future work includes testing LeanWAM in complex, dynamic real-world environments, scaling the Future Extractor with larger robot datasets and human videos, and integrating its fast control with a high-level agent for long-horizon planning and task decomposition.

{
    \small
    \bibliographystyle{ieeenat_fullname}
    \bibliography{main}
}

\clearpage
\onecolumn
\appendix

\begin{center}
{\Large\bfseries Appendix\par}
\end{center}

\section{Model Architecture}
\label{sec:supp_arch}

This section details the modules of both training stages (Sec.~\ref{sec:method}). Table~\ref{tab:supp_arch} summarizes the transformer modules that we introduce, and Table~\ref{tab:supp_mask} gives the attention mask of the policy. Sec.~\ref{sec:supp_training} lists which modules are trained in each stage. Blocks are indexed from 1 in all networks.

\subsection{Stage 1}
\label{sec:supp_arch_s1}

\paragraph{Future Extractor.}
The Future Extractor encodes the current frame and four future frames of the main camera (Table~\ref{tab:supp_setup}) with two frozen encoders. From the Wan2.1 VAE encoder~\cite{wan2025}, we take the 384-channel spatiotemporal features after its middle block, before the output head that maps them to latents. From DINOv3 ViT-L/16~\cite{dinov3}, we take the patch tokens of the penultimate block for each frame and normalize them with a LayerNorm without affine parameters. Both feature maps are pooled to a common spatial grid, and a linear layer maps the VAE features to the 1,024 channels of DINOv3. The tokens of the two encoders are then concatenated into one memory.

An 18-layer Q-Former~\cite{blip2} with 32 learnable queries reads this memory. Each layer applies pre-LN self-attention among the queries, cross-attention to the memory, and an FFN, each with a residual connection. Fixed 3D sinusoidal position embeddings over time, height, and width are added to the keys of the cross-attention but not to its values, which therefore carry only visual content. A final LayerNorm and a linear layer from 1,024 to 32 channels produce the code $z\in\mathbb{R}^{32\times32}$.

\paragraph{Conditioning and world tokens.}
A linear projection $P_g$ maps the code from 32 to 1,024 channels, and the result replaces the text embeddings in every cross-attention block of the frozen Cosmos-Predict2.5-2B~\cite{cosmos25}. The video model receives the clean latent of the current frame and the noised latent of the four future frames, which the VAE compresses into one latent frame. From block 18, we take the 2,048-channel hidden states at the future positions and project them linearly to 1,024 channels. The attentive pooler $R_\eta$ then aggregates them into the world tokens $\mathbf{w}\in\mathbb{R}^{32\times1024}$. It is a 3-layer Q-Former with 32 learnable queries, whose layers apply pre-LN self-attention, cross-attention with separate LayerNorms for the queries and the memory, and an FFN, followed by a final LayerNorm.

\paragraph{Action DiT.}
The action DiT is conditioned on the 64 tokens $[P_a(z);\,\mathbf{w}]$, where $P_a$ is a linear projection from 32 to 1,024 channels. Its input sequence contains a state token and the noisy action tokens. Its 16 blocks alternate between cross-attention to the conditioning tokens and self-attention over the input sequence, and all cross-attention blocks read the same conditioning tokens. The flow time modulates attention and FFN through AdaLN-Zero~\cite{dit}.

\paragraph{Alignment projectors.}
Two projectors map hidden states of the video model to 1,024-channel teacher features. The semantic projector reads block 9 and targets the penultimate-block patch tokens of DINOv3, normalized as in the Future Extractor but not pooled. The geometric projector reads block 14 and targets the patch tokens of the 14th global-attention block of VGGT-$\Omega$~\cite{vggtomega}, without additional normalization. Each projector consists of a spatial convolution and a two-layer MLP and keeps the spatial grid of the video model. The teacher features of the current frame are kept, and those of the four future frames are linearly interpolated in time to the future latent, so both the current and the future positions are aligned.

\subsection{Stage 2}
\label{sec:supp_arch_s2}

\paragraph{Vision-language backbone.}
The backbone keeps the full vision encoder of Qwen3.5-0.8B~\cite{qwen35} and its first 12 language layers, which form three groups of three Gated DeltaNet layers~\cite{gateddeltanet} followed by one gated attention layer. It reads the images of all cameras and the instruction, and the last-layer hidden states of all non-padding tokens form the context $\mathbf{L}$ with 1,024 channels. The backbone runs its original forward pass, so $\mathbf{L}$ is not updated by the experts.

\paragraph{Visual resampler.}
The visual resampler produces the current tokens $\mathbf{C}$ from a frozen DINOv3 ViT-L/16, which encodes each camera view separately. From blocks 13, 17, and 21, we take the class token, the register tokens, and the patch tokens. The features of each block pass through a separate learnable LayerNorm, and a linear layer maps their concatenation from 3,072 to 1,024 channels. The main camera keeps its full patch grid. For the other cameras, every $2\times2$ group of neighboring patches is folded into the channels and mapped from 4,096 to 1,024 channels by a second linear layer, which reduces the patch tokens by $4\times$ while keeping the class and register tokens. Both linear layers are shared across views. After adding a learnable embedding for each view, we concatenate the tokens of all views into one memory. A LayerNorm and a linear projection feed this memory to a 6-layer Q-Former with 64 learnable queries, built from pre-LN decoder blocks without a final LayerNorm, which outputs $\mathbf{C}\in\mathbb{R}^{64\times1024}$.

\paragraph{World and Action Experts.}
The World Expert processes $[\mathbf{C};\mathbf{F}]$, where $\mathbf{F}$ consists of 64 learnable future queries, 32 for the code and 32 for the world tokens. The Action Expert processes $\mathbf{A}$, which contains a state token, learnable register tokens, and the noisy action tokens. Input and output projections, sized to the state and action spaces of each benchmark (Table~\ref{tab:supp_setup}), map the states and actions to and from these tokens. The two experts have 12 layers each, with separate weights. At every layer, each expert computes the queries, keys, and values of its own tokens, and $\mathbf{L}$ is appended to the keys and values through a layer-specific LayerNorm and key and value projections. A single attention with the mask in Table~\ref{tab:supp_mask} then runs over all tokens, and its outputs return to the residual and FFN branches of each expert. The World Expert uses pre-LN blocks. The Action Expert uses AdaLN-Zero~\cite{dit}, in which the flow time produces a shift, a scale, and a residual gate for both attention and FFN.

Under this mask, future information reaches the actions only through $\mathbf{C}$, which reads $\mathbf{F}$ at every layer. Since neither $\mathbf{C}$ nor $\mathbf{F}$ attends to $\mathbf{A}$, the World Expert does not depend on the action noise or the integration step. Computing its keys and values once per query and reusing them across all Euler steps is therefore equivalent to recomputing them at every step.

\begin{table}[t]
\centering
\small
\setlength{\tabcolsep}{6pt}
\begin{tabular}{lccc}
\toprule
Module & Layers & Heads & Queries \\
\midrule
\multicolumn{4}{l}{\textit{Stage 1}} \\
Future Extractor Q-Former & 18 & 16 & 32 \\
Attentive pooler $R_\eta$ & 3 & 16 & 32 \\
Action DiT & 16 & 16 & -- \\
\midrule
\multicolumn{4}{l}{\textit{Stage 2}} \\
Visual resampler & 6 & 8 & 64 \\
World Expert & 12 & 16 & 64 \\
Action Expert & 12 & 16 & -- \\
\bottomrule
\end{tabular}
\caption{\textbf{Transformer modules introduced by LeanWAM.} All have a hidden width of 1,024 and an FFN width of 4,096. Queries are learnable tokens, which form the future queries $\mathbf{F}$ in the World Expert.}
\label{tab:supp_arch}
\end{table}

\begin{table}[t]
\centering
\small
\setlength{\tabcolsep}{8pt}
\begin{tabular}{lcccc}
\toprule
 & \multicolumn{4}{c}{Keys and values} \\
\cmidrule(lr){2-5}
Queries & $\mathbf{L}$ & $\mathbf{C}$ & $\mathbf{F}$ & $\mathbf{A}$ \\
\midrule
$\mathbf{C}$ (World Expert) & \cmark & \cmark & \cmark & \xmark \\
$\mathbf{F}$ (World Expert) & \cmark & \cmark & \cmark & \xmark \\
$\mathbf{A}$ (Action Expert) & \xmark & \cmark & \xmark & \cmark \\
\bottomrule
\end{tabular}
\caption{\textbf{Attention mask of each expert layer.} $\mathbf{L}$: context from the backbone, used only as keys and values. $\mathbf{C}$: current tokens. $\mathbf{F}$: future queries. $\mathbf{A}$: state, learnable register tokens, and noisy action tokens.}
\label{tab:supp_mask}
\end{table}

\paragraph{Prediction heads.}
At the last layer of the World Expert, the first 32 future queries predict the code through a LayerNorm and a linear layer from 1,024 to 32 channels, and the other 32 predict the world tokens through a separate LayerNorm and a linear layer from 1,024 to 1,024 channels. The token-wise LayerNorm without affine parameters is applied only to the distillation targets, and the video model is always conditioned on the unnormalized code. The heads are used only for training, so the deployed policy consists of the other modules in this subsection (Sec.~\ref{sec:supp_efficiency}).

\paragraph{Initialization.}
Both experts are randomly initialized, and the Action Expert does not inherit the weights of the Stage-1 action DiT. In the World Expert, the output projections of attention and FFN are scaled down with layer depth. The AdaLN-Zero modulation of the Action Expert is initialized to zero, so each of its blocks starts as the identity function~\cite{dit}.

\section{Details of the Probing Study}
\label{sec:supp_probe}

This section gives the full protocol of the probing study in Sec.~\ref{sec:probe}.

\paragraph{Data.}
We use the LIBERO demonstrations~\cite{libero}. For each of three random seeds, we sample 400 episodes for training and 100 different episodes for evaluation, covering all tasks uniformly. Because the split is made at the level of episodes, no trajectory, and hence no overlapping action chunk, appears in both sets. The three seeds use different splits, and all variants are trained and evaluated on the same splits. At each time step, the two camera views are concatenated side by side into a $224\times448$ image before being encoded. A sample consists of the current frame $I_t$, the action chunk $\mathbf{a}_t=(a_t,\ldots,a_{t+H-1})$ with $H=32$, and the future frames within the same horizon. The two-frame setting observes $I_{t+H}$, the three-frame setting observes $\{I_{t+H/2},I_{t+H}\}$, and the five-frame setting observes $\{I_{t+H/4},I_{t+H/2},I_{t+3H/4},I_{t+H}\}$.

\paragraph{Features.}
All feature extractors are frozen.
\textbf{D:} DINOv3 ViT-L/16 with 300M parameters~\cite{dinov3} encodes each frame independently, and we take the patch tokens of its last layer.
\textbf{W:} the Wan2.1 VAE encoder of Cosmos-Predict2.5~\cite{wan2025,cosmos25} encodes the frame sequence, and we take its output latents. The encoder maps the first frame to one latent and each subsequent group of four frames to another, so we repeat the future frames to fill this four-frame window: a single future frame is repeated four times, and each of two future frames is repeated twice.
\textbf{F and T:} the same latents are fed to Cosmos-Predict2.5-2B~\cite{cosmos25}, with the current latent kept clean and the future latent noised to level $\tau$, where $\tau=1$ corresponds to pure noise. We take the hidden states of block $k$ at all positions, current and future. Both keep the original text conditioning of Cosmos-Predict2.5-2B, whose text embeddings come from Cosmos-Reason1-7B~\cite{cosmosreason1}. The default setting is $k=18$ and $\tau=0.75$. The variants T$^*$ use $k\in\{12,24\}$ with $\tau=0.75$, or $\tau\in\{0.25,1.0\}$ with $k=18$. F uses the pretrained weights. T uses the weights obtained by training the video model together with an action DiT on the 400 training episodes, under a flow-matching video loss and action loss, as in a standard WAM. T is never trained on the 100 evaluation episodes, so these episodes are held out for both T and the probe.

\paragraph{Probe.}
Table~\ref{tab:probe} lists the architecture. Each source is mapped to width 512 by its own linear layer. When several sources are combined, each receives a learnable source embedding, and the tokens of all frames and sources are concatenated into one sequence. The aggregator has $N$ learnable queries (default $N=32$) and two blocks, each consisting of cross-attention from the queries to the feature tokens followed by an FFN. The action decoder has $H$ learnable action queries, one per time step, and two blocks, each consisting of self-attention among the action queries, cross-attention to the $N$ aggregated tokens, and an FFN. A final layer normalization and a linear layer map each action query to an action. We use an attentive probe~\cite{vjepa} rather than a linear one because the sources differ in the number, width, and layout of their tokens. The aggregator and action decoder use the same width and depth across variants.

\begin{table}[t]
\centering
\small
\begin{tabular}{@{}ll@{}}
\toprule
Component & Configuration \\
\midrule
Input adapter & Linear to 512, one per source \\
Source embedding & learnable, one per source$^\dagger$ \\
Aggregator queries & $N\times512$, default $N=32$ \\
Aggregator & $2\times$[cross-attn., FFN] \\
Action queries & $H\times512$, one per time step \\
Action decoder & $2\times$[self-attn., cross-attn., FFN] \\
Attention & width 512, 8 heads \\
FFN & $512\to2048\to512$, GELU \\
Normalization & pre-LN, residual per sublayer \\
Output head & LayerNorm, Linear to action dim. \\
Dropout & 0 \\
\midrule
Optimizer & AdamW, weight decay 0 \\
Learning rate & $2\times10^{-4}$, constant, no warmup \\
Batch size & 256 \\
Training & 2 epochs over 400 episodes \\
\bottomrule
\end{tabular}
\caption{\textbf{Configuration of the attentive probe.} Only the probe is trained, and all feature extractors stay frozen. $^\dagger$Used only when several sources are combined.}
\label{tab:probe}
\end{table}

\paragraph{Training and evaluation.}
Only the probe is trained. Actions are normalized, and the probe minimizes the mean squared error between the predicted and ground-truth action chunks. We train for two epochs over the 400 training episodes with AdamW, a constant learning rate of $2\times10^{-4}$ without warmup, no weight decay, and a batch size of 256. All variants use the same settings. We evaluate on all action chunks of the 100 held-out episodes. For each of the $d_a=7$ action dimensions, we compute the coefficient of determination over all chunks and time steps, and we report the unweighted mean over dimensions:
\begin{equation}
R^2=\frac{1}{d_a}\sum_{d=1}^{d_a}\left(1-\frac{\sum_{n}\sum_{h=1}^{H}\big(\hat{a}_{n,h,d}-a_{n,h,d}\big)^2}{\sum_{n}\sum_{h=1}^{H}\big(a_{n,h,d}-\bar{a}_{d}\big)^2}\right),
\label{eq:r2}
\end{equation}
where $n$ indexes evaluation chunks, $a_{n,h,d}$ and $\hat{a}_{n,h,d}$ are the ground-truth and predicted values of dimension $d$ at step $h$, and $\bar{a}_{d}$ is the mean of dimension $d$ over all chunks and steps. We report the mean and standard deviation over the three seeds. The error bars in Figure~\ref{fig:probe} show $\pm1$ standard deviation, not the standard error or a confidence interval.

\section{Experimental Details}
\label{sec:supp_exp}

This section details the protocols and training settings summarized in Sec.~\ref{sec:exp_setup}, and reports per-task results on RoboTwin~2.0. Table~\ref{tab:supp_setup} summarizes the setup on each benchmark.

\begin{table}[t]
\centering
\footnotesize
\setlength{\tabcolsep}{4pt}
\begin{tabular}{lcc}
\toprule
 & LIBERO / LIBERO-Plus & RoboTwin~2.0 \\
\midrule
Robot & Franka Panda & Aloha-AgileX \\
Training tasks & 40 & 10 \\
Demonstrations & 1,693 & 5,338 \\
Training frames & 273,465 & 1,161,718 \\
Cameras & third-person, wrist & head, two wrists \\
Image size (W$\times$H) & 224$\times$224 & 320$\times$256 \\
State & 8-D, EE pose & 14-D, joints \\
Action & 7-D, EE delta & 14-D, joint targets \\
Chunk / executed & 16 / 10 & 32 / 16 \\
Stage-1 horizon & 32 steps & 32 steps \\
Test episodes & 2,000 / 10,030 & 1,000 / 1,000 \\
\bottomrule
\end{tabular}
\caption{\textbf{Setup on each benchmark.}}
\label{tab:supp_setup}
\end{table}

\subsection{Benchmarks and Protocols}
\label{sec:supp_protocols}

\paragraph{LIBERO.}
LIBERO~\cite{libero} has four suites of 10 language-conditioned tasks, performed by a simulated Franka Panda arm. LIBERO-Spatial places the same objects in different layouts, LIBERO-Object varies the objects to be manipulated, LIBERO-Goal varies the goal in a fixed scene, and LIBERO-Long, also known as LIBERO-10, contains long-horizon tasks that chain several subtasks. Each task has 50 human demonstrations. We use the version released by OpenVLA~\cite{openvla}, which replays the demonstrations to render $256\times256$ images, removes no-op actions, and discards demonstrations that fail on replay. This leaves 1,693 demonstrations with 273,465 frames, on which we train a single model for all four suites.

The policy receives the instruction, the third-person and wrist images resized to $224\times224$, and an 8-dimensional proprioceptive state: the end-effector position, its orientation in axis-angle form, and the positions of the two gripper fingers. It predicts chunks of 16 actions, each a 7-dimensional command of end-effector translation and rotation deltas and a gripper action, and executes the first 10 before predicting again. Stage~1 uses only the third-person camera, with clips of the current frame and four future frames over the next 32 steps. This horizon is set independently of the Stage-2 chunk length, so the distilled targets describe a longer future than each predicted chunk.

Following the standard protocol, we evaluate each task over 50 episodes from the initial states provided by the benchmark, giving 500 episodes per suite and 2,000 in total. Each episode begins with 10 idle steps that let the objects settle, after which the policy acts for at most 220, 280, 300, and 520 steps on Spatial, Object, Goal, and Long, respectively. An episode is successful if the task is completed within this limit.

\paragraph{LIBERO-Plus.}
LIBERO-Plus~\cite{liberoplus} evaluates policies trained on LIBERO under controlled perturbations along seven dimensions:
\begin{itemize}[leftmargin=*,itemsep=0pt,topsep=2pt]
\item \textbf{Camera}: the pose and field of view of the third-person camera.
\item \textbf{Robot}: the initial pose of the arm.
\item \textbf{Language}: instructions rewritten with richer and more complex language.
\item \textbf{Light}: the intensity, direction, and color of the illumination, and the resulting shadows.
\item \textbf{Background}: the textures and materials of the table and the scene.
\item \textbf{Noise}: photometric distortions of the input images, such as jitter and Gaussian blur.
\item \textbf{Layout}: added distractor objects and shifted target positions.
\end{itemize}
The benchmark contains 10,030 perturbed task instances, each evaluated once. We evaluate the LIBERO checkpoint of LeanWAM with the same inputs, chunking, integration steps, and step limits as on LIBERO, so the two benchmarks differ only in the perturbations and in the number of episodes per task. For all methods in Table~\ref{tab:liberoplus}, task-specific training uses only LIBERO demonstrations, and no method is trained on LIBERO-Plus data. Following the benchmark leaderboard, we report the success rate for each dimension and over all 10,030 episodes (Total).

\paragraph{RoboTwin 2.0.}
RoboTwin~2.0~\cite{robotwin2} contains 50 bimanual tasks, which we run with the Aloha-AgileX dual-arm robot as in its benchmark evaluation. Each task is evaluated in two settings, called Easy and Hard in the original paper and clean and randomized in ours. The clean setting uses scenes without domain randomization, while the randomized setting adds cluttered distractors and randomizes background textures, lighting, and table height. We use the 10 tasks in Table~\ref{tab:supp_robotwin}. The official dataset provides 50 clean and 500 randomized demonstrations per task. We discard demonstrations whose arm-joint actions leave the valid range and trim static frames, which leaves 5,338 demonstrations with 1,161,718 frames, and train a single model on all 10 tasks.

The policy receives the instruction, the images of the head camera and the two wrist cameras resized to $320\times256$ (width $\times$ height), and a 14-dimensional state with the six joint positions of each arm and the two gripper states. It predicts chunks of 32 actions, each a 14-dimensional vector of absolute joint and gripper targets, and executes the first 16. Unlike the end-effector deltas on LIBERO, these actions are absolute targets in joint space. Stage~1 uses only the head camera, with clips of the current frame and four future frames over the next 32 steps. We evaluate each task over 100 episodes in each setting.

\subsection{Training and Inference Details}
\label{sec:supp_training}

Both stages of LeanWAM are trained on a single node with 8 NVIDIA A100 GPUs (80\,GB), using DeepSpeed ZeRO-2~\cite{zero} and BF16 mixed precision, with the action, video, alignment, and distillation losses computed in FP32. We use AdamW~\cite{adamw} with $\beta_1=0.9$, $\beta_2=0.99$, and $\epsilon=10^{-8}$, a linear warmup followed by cosine decay to a minimum learning rate, and gradient clipping at a global norm of 1.0. Table~\ref{tab:supp_hparams} lists the settings that differ across stages and benchmarks.

\begin{table*}[t]
\centering
\small
\setlength{\tabcolsep}{7pt}
\begin{tabular}{lcccc}
\toprule
 & \multicolumn{2}{c}{Stage~1} & \multicolumn{2}{c}{Stage~2} \\
\cmidrule(lr){2-3}\cmidrule(lr){4-5}
 & LIBERO & RoboTwin~2.0 & LIBERO & RoboTwin~2.0 \\
\midrule
Epochs & 30 & 10 & 30 & 15 \\
Global batch size & 256 & 256 & 1,024 & 256 \\
Warmup (fraction of steps) & 2\% & 2\% & 5\% & 2\% \\
Weight decay & 0 & 0 & $10^{-8}$ & $10^{-8}$ \\
Learning rate, Stage-1 modules & $3{\times}10^{-4}\rightarrow2{\times}10^{-5}$ & $2{\times}10^{-4}\rightarrow2{\times}10^{-5}$ & -- & -- \\
Learning rate, Qwen3.5 backbone & -- & -- & $10^{-4}\rightarrow5{\times}10^{-7}$ & $10^{-4}\rightarrow5{\times}10^{-7}$ \\
Learning rate, other modules & -- & -- & $2{\times}10^{-4}\rightarrow10^{-5}$ & $2{\times}10^{-4}\rightarrow3{\times}10^{-5}$ \\
Noise level $\tau$ of the video model & \multicolumn{2}{c}{$5u/(1+4u)$, random} & \multicolumn{2}{c}{$\tau^\star=0.75$, fixed} \\
Auxiliary loss weights & \multicolumn{2}{c}{$\lambda_{\mathrm{sem}}=\lambda_{\mathrm{geo}}=0.1$} & \multicolumn{2}{c}{0.1 (code), 0.1 (world)} \\
\bottomrule
\end{tabular}
\caption{\textbf{Training hyperparameters.} Learning rates are given as peak $\rightarrow$ minimum.}
\label{tab:supp_hparams}
\end{table*}

\paragraph{Stage 1.}
Stage~1 trains the Future Extractor, the attentive pooler, the action DiT, and the projections for conditioning and alignment. The video model, its VAE encoder, and DINOv3 are frozen, but the gradients of the video and action losses still flow through the video model to the code. The current latent is kept clean, and the future latents are noised to the level $\tau=5u/(1+4u)$, where $u$ is drawn uniformly from 1,000 discrete levels. This shift places more samples at high noise levels. The world features that condition the action DiT come from the same forward pass and thus share this noise level.

\paragraph{Stage 2.}
Stage~2 fine-tunes all parameters of the first 12 layers of Qwen3.5-0.8B, together with the World Expert, the Action Expert, the visual resampler, and the prediction heads, while DINOv3 and all Stage-1 modules stay frozen. The backbone uses a lower learning rate than the other modules. We precompute the distillation targets once. For the world tokens, the video model runs at the fixed noise level $\tau^\star=0.75$ with a fixed noise sample per clip, so the targets are the same in every epoch.

\paragraph{Inference.}
The Action Expert generates each action chunk by integrating its velocity field over 10 Euler steps. The backbone and DINOv3 run in BF16. The experts use BF16 autocast for matrix operations, while normalization and the action noise and integration state remain in FP32 (Sec.~\ref{sec:supp_efficiency}).

\subsection{Baselines}
\label{sec:supp_baselines}

In the tables of the main paper, Size counts the parameters of the deployed model. PT marks methods whose policy is pretrained to predict actions or latent actions on large-scale robot or human manipulation data before training on the benchmark. LeanWAM builds on pretrained vision, language, and video models, none of which is pretrained on embodied action data.

On LIBERO and LIBERO-Plus, task-specific training uses only LIBERO demonstrations. Methods marked PT may additionally use large-scale embodied pretraining. On RoboTwin~2.0, the baselines are trained on all 50 tasks in their papers, and we average their reported per-task success rates over our 10 tasks. LeanWAM is trained on these 10 tasks only. The two protocols thus differ: the baselines learn from more tasks and demonstrations, while LeanWAM is trained on fewer tasks.

\subsection{Per-Task Results on RoboTwin 2.0}
\label{sec:supp_robotwin}

Table~\ref{tab:supp_robotwin} reports the success rate of LeanWAM on each task. LeanWAM succeeds in at least 92\% of the clean episodes on eight of the ten tasks, and its success rate changes by at most 3 points under randomization on eight tasks. Turn Switch and Pick Diverse Bottles are the most challenging tasks, and Turn Switch also drops the most under randomization, from 83\% to 74\%.

\begin{table}[t]
\centering
\small
\setlength{\tabcolsep}{10pt}
\begin{tabular}{lcc}
\toprule
Task & Clean & Randomized \\
\midrule
Adjust Bottle & 100 & 99 \\
Handover Block & 96 & 93 \\
Open Laptop & 98 & 99 \\
Pick Diverse Bottles & 85 & 85 \\
Place Object Basket & 92 & 94 \\
Press Stapler & 96 & 93 \\
Stack Blocks Two & 98 & 100 \\
Stack Bowls Three & 93 & 88 \\
Stack Bowls Two & 98 & 98 \\
Turn Switch & 83 & 74 \\
\midrule
Average & 93.9 & 92.3 \\
\bottomrule
\end{tabular}
\caption{\textbf{Per-task results of LeanWAM on RoboTwin 2.0.} Success rates (\%) over 100 episodes per task in each setting.}
\label{tab:supp_robotwin}
\end{table}

\subsection{Ablation Details}
\label{sec:supp_ablation}
 
The ablations in Sec.~\ref{sec:ablation} follow the LIBERO-Plus protocol in Sec.~\ref{sec:supp_protocols}. Each variant is trained on LIBERO with the same data, schedule, and hyperparameters as LeanWAM and is evaluated on all 10,030 episodes of LIBERO-Plus. A Stage-1 variant changes the Stage-1 model, and Stage~2 is retrained with it as the teacher. A Stage-2 variant reuses the Stage-1 model of LeanWAM. Each variant is trained once. Under an independent Bernoulli approximation, the evaluation standard error is about 0.4 percentage points.
 
\paragraph{Stage-1 variants.}
\begin{itemize}[leftmargin=*,itemsep=1pt,topsep=2pt]
\item \textbf{$-$ action supervision} removes the action DiT and its loss, so the Future Extractor learns only from the video and alignment losses through the frozen video model. Without the action DiT, there is no attentive pooler and hence no world tokens, and Stage~2 is trained with the action loss and the code distillation loss.
\item \textbf{$-$ video model} removes the video model, together with the alignment losses and the world tokens. The Future Extractor learns only from the action loss, through an action DiT conditioned on the projected code alone. Stage~2 again distills only the code.
\item \textbf{$-$ semantic and geometric alignment} removes $\mathcal{L}_{\mathrm{sem}}$ and $\mathcal{L}_{\mathrm{geo}}$ and keeps the action and video losses. Stage~2 distills both targets, as in LeanWAM.
\item \textbf{Frozen $\rightarrow$ trainable video model} unfreezes Cosmos-Predict2.5-2B and updates it with the action, video, and alignment losses at a constant learning rate of $10^{-5}$. All other modules and settings are unchanged.
\item \textbf{Alignment $\rightarrow$ prediction heads} replaces the alignment losses with two output heads that predict future DINOv3 features and depth, as described next.
\end{itemize}
 
\paragraph{Prediction heads.}
This variant follows WAMs that add semantic or geometric targets to the predicted future~\cite{flexpi,dreamwam}. The video model predicts the velocity of the future RGB latent from the hidden states of its last block, through adaptive layer normalization, a linear layer, and unpatchifying. We add two heads of the same form next to it, each with its own adaptive layer normalization, a linear layer from width 2048 to the target dimension, and unpatchifying. Both heads are randomly initialized and read the same hidden states as the RGB head. The input of the video model is unchanged: the clean current latent and the noisy future latent.
 
The semantic head predicts DINOv3 features~\cite{dinov3}. Since the Wan VAE~\cite{wan2025} compresses the four future frames into one latent, we take the target from the last of them: the patch features of the frozen DINOv3 ViT-L/16 after its second-to-last block, followed by a layer normalization without affine parameters. These features have 1,024 channels and the same spatial grid as the tokens of the video model, so the head maps 2048 channels to 1,024 without any resampling. The head regresses the features directly, without noise, under a mean squared error.
 
The geometric head predicts depth. We compute pseudo-depth maps of the future frames with VGGT-$\Omega$~\cite{vggtomega}, clip them to a fixed range estimated on the training set, and apply a logarithmic normalization, using the same range for all videos of a camera. The normalized maps are replicated to three channels and encoded by the frozen Wan VAE encoder, which compresses the four future depth maps into one latent $\mathbf{z}_{\mathrm{d}}$. Like the RGB head, the depth head is trained with flow matching and predicts the velocity $\boldsymbol{\epsilon}-\mathbf{z}_{\mathrm{d}}$. Both head losses have a weight of 1.0.
 
The rest of Stage~1 is unchanged. The attentive pooler reads the world tokens from the same block, the action DiT is conditioned on the code and the world tokens, and both serve as targets in Stage~2.
 
\paragraph{Stage-2 variants.}
\begin{itemize}[leftmargin=*,itemsep=1pt,topsep=2pt]
\item \textbf{$-$ code distillation}, \textbf{$-$ world distillation}, and \textbf{$-$ code and world distillation} remove the corresponding distillation losses from the Stage-2 objective. The architecture, including the future queries, is unchanged, so these variants have the same inference cost as LeanWAM.
\item \textbf{DINOv3 tokens $\rightarrow$ learnable tokens} replaces the 64 current tokens, produced by DINOv3 and the visual resampler, with 64 learnable tokens. DINOv3 and the resampler are removed from both training and inference, so the policy observes the images only through the vision-language backbone.
\end{itemize}

\section{Inference Efficiency and Training Cost}
\label{sec:supp_efficiency}

\paragraph{Latency measurement.}
We time the complete prediction of one action chunk on a single NVIDIA A100-SXM4-80GB GPU with batch size 1 and BF16 mixed precision, using the same GPU for all methods. Each method answers 1,000 queries on the ten LIBERO-Long tasks, 100 per task. We synchronize the GPU immediately before and after the neural computation of each query and measure the wall-clock time in between, which includes every network component that the query requires. We exclude input preprocessing, host-to-device transfer, simulation and rendering, communication, output post-processing, and robot execution. Warmup, compilation, and graph capture are excluded as well: each method runs 30 warmup queries before timing and three untimed queries whenever the task changes. We report the mean and standard deviation over all timed queries without discarding slow samples. Each baseline is timed with its official implementation, keeping its native input resolution and prediction horizon. For $\pi_{0.5}$, this is the JAX implementation in openpi\footnote{\url{https://github.com/Physical-Intelligence/openpi}}.

\paragraph{LeanWAM at inference.}
LeanWAM receives two $224\times224$ images, the instruction, and the 8-dimensional state, and predicts a chunk of 16 actions, of which 10 are executed before the next query. For each query, the backbone and the World Expert run once and the Action Expert runs ten Euler steps. The deployed policy contains no video model and no other Stage-1 module. The backbone and DINOv3 run in BF16, the experts run their matrix operations under BF16 autocast, and the normalization layers and the noise and integration state of the actions stay in FP32. We implement this policy in PyTorch~\cite{pytorch}, which we also use for training, and in JAX, which gives the latency reported in the main paper. Both implementations load the same Stage-2 checkpoint.

\paragraph{PyTorch implementation.}
The PyTorch implementation uses four optimizations that do not change what the model computes:
\begin{enumerate}[leftmargin=*,itemsep=1pt,topsep=2pt]
\item \textbf{Parallel visual encoding.} The backbone runs on one CUDA stream while DINOv3 and the visual resampler run on another, and the two branches join before the experts.
\item \textbf{Compact action attention.} Since the action tokens attend only to the current tokens $\mathbf{C}$ and to themselves (Sec.~\ref{sec:stage2}), we pass only the keys and values of $[\mathbf{C},\mathbf{A}]$ to their attention instead of masking all other tokens. The World Expert is computed in full.
\item \textbf{Compiled denoising.} A static-shape denoising step of the Action Expert is compiled with TorchInductor. Masks, position embeddings, and the state conditioning are built once per query, and the keys and values of $\mathbf{C}$ are computed once and reused across the ten steps.
\item \textbf{CUDA graphs.} DINOv3 with the visual resampler, and the World and Action Experts, are captured as two CUDA graphs and replayed for every query, while the backbone runs eagerly. Each query uses its own observation, state, and action noise, and nothing is cached across queries.
\end{enumerate}

\paragraph{JAX implementation.}
We obtain a JAX~\cite{jax2018github} implementation with torchax~\cite{torchax}, a PyTorch frontend for JAX, which runs our PyTorch modules as JAX programs that XLA compiles. This implementation therefore reuses the network definitions of the PyTorch one. Each query calls four functions compiled with \texttt{jax.jit}: the vision encoder of Qwen3.5, DINOv3 with the visual resampler, the language model of Qwen3.5, and the World and Action Experts. The last function runs all ten Euler steps, with the same compact action attention and the same reuse of the keys and values of $\mathbf{C}$ as the PyTorch implementation. We use neither separate CUDA streams nor explicit CUDA graphs, and we precompute only the image-grid indices and the rotary-embedding constants, which are identical for all queries.

Nine of the 12 language layers that we keep from Qwen3.5 use Gated DeltaNet~\cite{gateddeltanet}, a linear attention mechanism based on the gated delta rule. We implement its chunkwise form natively in JAX, with chunks of 64 tokens, a triangular solve within each chunk, and a \texttt{lax.scan} over chunks that keeps the recurrent state in FP32. We align the autocast rules of torchax with those of PyTorch, so the two implementations use the same mixed precision. Softmax attention uses the fused cuDNN kernel when its inputs are in BF16 and its head dimension is at most 128, and the XLA implementation otherwise. We time the JAX implementation on 1,000 queries, 100 per task, synchronizing the GPU with \texttt{jax.block\_until\_ready}.

\begin{table}[t]
\centering
\small
\setlength{\tabcolsep}{8pt}
\begin{tabular}{lcc}
\toprule
Method & Latency (ms) & Relative \\
\midrule
OpenVLA-OFT~\cite{openvlaoft} & $104.2\pm1.2$ & 4.2$\times$ \\
$\pi_{0.5}$~\cite{pi05} & $65.8\pm\mathrm{0.2}$ & 2.6$\times$ \\
GR00T N1.7~\cite{gr00tn1} & $267.5\pm15.5$ & 10.7$\times$ \\
VLAct~\cite{vlact} & $531.4\pm6.8$ & 21.2$\times$ \\
\midrule
Fast-WAM~\cite{fastwam} & $627.1\pm42.9$ & 25.0$\times$ \\
OpenWAM-$\alpha$~\cite{openwam} & $1040.3\pm129.3$ & 41.5$\times$ \\
Cosmos Policy~\cite{cosmospolicy} & $788.2\pm74.7$ & 31.4$\times$ \\
ImageWAM~\cite{imagewam} & $307.6\pm23.3$ & 12.3$\times$ \\
LingBot-VA~\cite{lingbotva} & $1203.4\pm19.7$ & 47.9$\times$ \\
\midrule
\textbf{LeanWAM} (PyTorch) & $63.1\pm3.4$ & 2.5$\times$ \\
\textbf{LeanWAM} (JAX) & $\mathbf{25.1\pm0.7}$ & 1.0$\times$ \\
\bottomrule
\end{tabular}
\caption{\textbf{Inference latency.} Mean and standard deviation of the time to predict one action chunk on an A100 GPU with batch size 1, over 1,000 queries. Relative: latency divided by that of LeanWAM (JAX). VLAs are listed above WAMs.}
\label{tab:supp_latency}
\end{table}

\paragraph{Latency results.}
Table~\ref{tab:supp_latency} lists all methods we timed, including GR00T N1.7 and LingBot-VA, which appear in the LIBERO and RoboTwin~2.0 comparisons. With the JAX implementation, LeanWAM takes $25.1\pm0.7$\,ms per action chunk, which corresponds to 39.8 queries per second. It is 2.6$\times$ faster than $\pi_{0.5}$, the fastest baseline, 4.2$\times$ faster than OpenVLA-OFT, and 12.3$\times$ to 47.9$\times$ faster than the five WAMs. Both LeanWAM and $\pi_{0.5}$ are timed in JAX, so this comparison uses the same framework. Over 1,000 queries, the PyTorch implementation takes $63.1\pm3.4$\,ms, with a median of 61.3\,ms and a 95th percentile of 69.3\,ms. JAX is 2.5$\times$ faster under this protocol. The PyTorch implementation is also faster than every baseline measured here.

\paragraph{Memory and parameters.}
We measure memory with the PyTorch implementation, in a separate process with batch size 1, over 200 queries on the same ten tasks after warmup. The PyTorch allocator holds at most 3.76\,GiB, including the model weights and the CUDA-graph buffers. NVML reports at most 4.58\,GiB for the process, which also includes the CUDA context and libraries. Neither number includes the simulator. The deployed policy has 1.431B parameters: 604M in the truncated Qwen3.5 backbone, 303M in the frozen DINOv3 encoder, and 524M in the World and Action Experts and the visual resampler. The count excludes the Stage-1 modules and the teachers, which the deployed policy does not contain.

\begin{table}[t]
\centering
\small
\setlength{\tabcolsep}{5pt}
\begin{tabular}{lccccc}
\toprule
 & GPUs & Epochs & Updates & Hours & GPU-hours \\
\midrule
Stage~1 & 8 & 30 & 32,040 & 18.1 & 144.5 \\
Stage~2 & 8 & 30 & 8,010 & 12.4 & 99.0 \\
\midrule
Total & & & & 30.4 & 243.5 \\
\bottomrule
\end{tabular}
\caption{\textbf{Training cost on LIBERO.} Wall-clock hours and A100 GPU-hours of the two stages.}
\label{tab:supp_train_cost}
\end{table}

\paragraph{Training cost.}
Table~\ref{tab:supp_train_cost} reports the training cost on LIBERO. Both stages run on a single node with eight A100 GPUs and take 30.4 hours in total. Stage~1 runs the frozen video model online and reads its alignment targets from a cache. Stage~2 reads cached distillation targets and does not load the Stage-1 model. The costs exclude the one-time computation of these caches, the pretraining of the backbones, and evaluation. Stage-2 training peaks at about 18.9\,GiB of GPU memory with 128 samples per GPU.

\section{Additional Query Attention Visualizations}
\label{sec:supp_attn}

\begin{figure}[!htbp]
    \centering
    \includegraphics[width=0.75\linewidth]{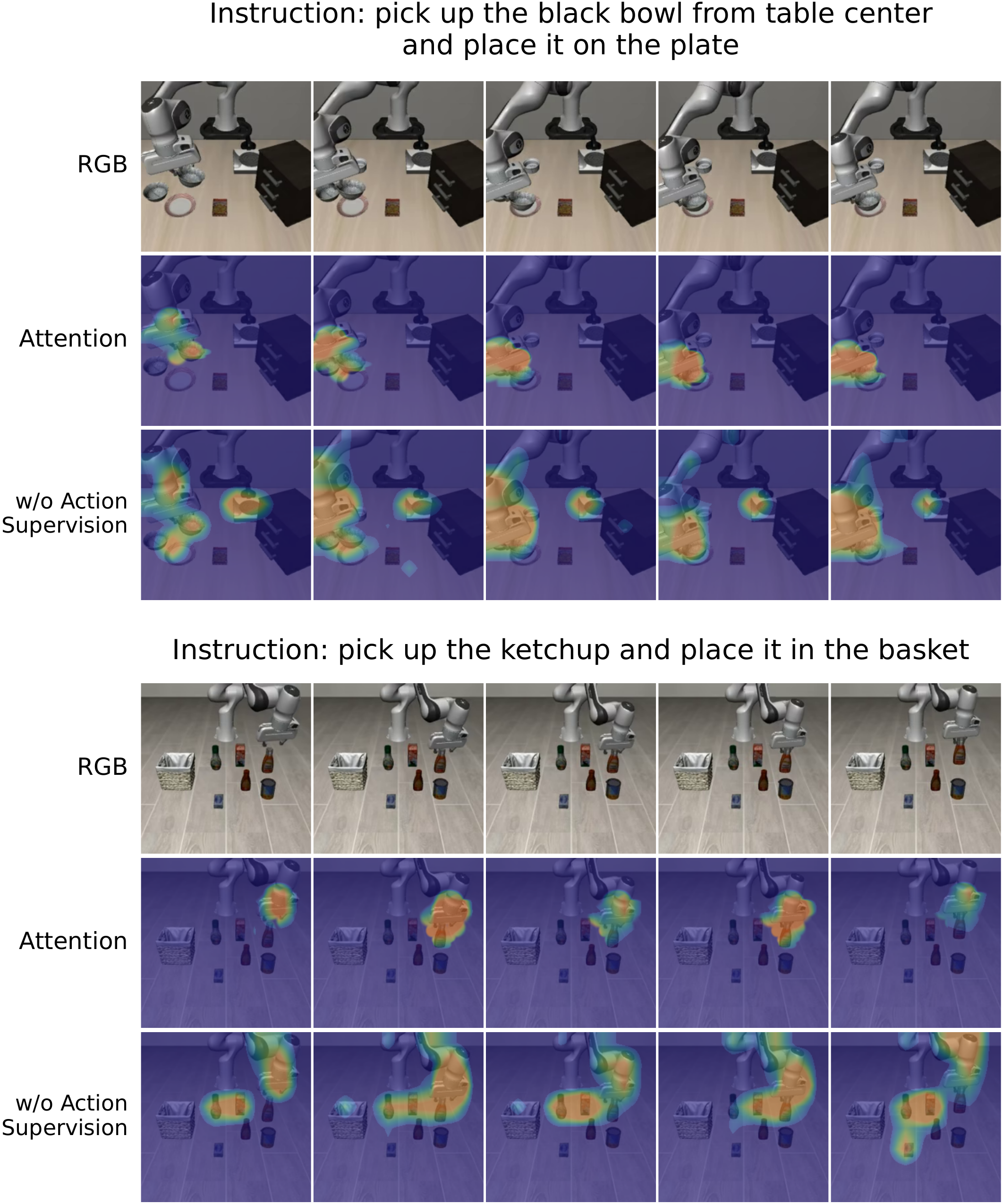}
    \caption{\textbf{Query attention of the Future Extractor with and without action supervision.} For each task, the rows show the five frames of one input clip (RGB), the attention of our Future Extractor (Attention), and the attention of a Future Extractor trained without the Stage-1 action loss (w/o Action Supervision).}
    \label{fig:supp_attn}
\end{figure}

Sec.~\ref{sec:analysis} visualizes where the queries of the Future Extractor attend in one LIBERO demonstration. Figure~\ref{fig:supp_attn} shows two more tasks, from LIBERO-Spatial and LIBERO-Object, and compares our Future Extractor with one trained without the Stage-1 action loss.

\paragraph{Visualization.}
The Future Extractor reads the current frame and four future frames through an 18-layer Q-Former with 32 learnable queries (Sec.~\ref{sec:supp_arch_s1}). In every block, the queries cross-attend to a memory of VAE and DINOv3 tokens, which gives one attention matrix per head. We keep the weights on the DINOv3 tokens, which retain a separate spatial map for each frame, and average them over all heads, all 32 queries, and the last 8 of the 18 blocks. The averaged weights of each frame are reshaped into a spatial map, resized to the image, and overlaid on the frame. The variant without action supervision is trained with the same Stage-1 objective except the action loss and is visualized in the same way. The visualization in Figure~\ref{fig:attn} follows the same procedure.

\paragraph{Results.}
With action supervision, the attention concentrates on the gripper and the target object throughout the clip. In the first task, it covers the gripper and the black bowl that the robot carries toward the plate. In the second, it follows the gripper as the robot grasps and lifts the ketchup bottle. The rest of the scene receives little attention. Without action supervision, the attention spreads over the robot arm and over objects that the task does not involve: an object on the stove in the first task, and the bottles and boxes next to the ketchup in the second. In these examples, action supervision is associated with more concentrated attention on the end effector and manipulated object, consistent with the behavior analysis in Sec.~\ref{sec:analysis}.

\end{document}